\documentclass[journal,twoside,web]{ieeecolor}
\usepackage{tmi}
\usepackage{cite}
\usepackage{amsmath,amssymb,amsfonts}
\usepackage{algorithmic}
\usepackage{graphicx}
\usepackage{algorithm,algorithmic}
\usepackage{hyperref}
\hypersetup{hidelinks=true}
\usepackage{textcomp}
\usepackage[T1]{fontenc}
\usepackage{adjustbox}
\usepackage{booktabs, threeparttable}
\usepackage{siunitx}
\usepackage{multirow} 
\usepackage{pifont}
 \usepackage{url}
\usepackage[figuresright]{rotating}
\usepackage{subcaption}
\allowdisplaybreaks
\usepackage[table]{xcolor}
\definecolor{colA}{RGB}{200,240,200}
\definecolor{colB}{RGB}{255,210,210}
\definecolor{colC}{RGB}{210,225,255}
\definecolor{colD}{RGB}{255,250,205}
\definecolor{colE}{RGB}{235,215,255}
\definecolor{colF}{RGB}{255,235,210}
\definecolor{light_grey}{RGB}{200, 200, 200}
\newcommand{\epochs}[1]{\textcolor{black!80}{#1}}
\newcommand{\iters}[1]{\textcolor{black!50}{#1}}
\newcommand{\cmark}{\textcolor{green!70!black}{\checkmark}}
\newcommand{\xmark}{\textcolor{red!70!black}{\texttimes}}
\let\oldding\ding
\renewcommand{\ding}[2][1]{\scalebox{#1}{\oldding{#2}}}
\newcommand{\numone}{\textcolor{red!90!black}{\ding[1.2]{182}}}
\newcommand{\numtwo}{\textcolor{orange!90!black}{\ding[1.2]{183}}}
\newcommand{\numthree}{\textcolor{yellow!60!black}{\ding[1.2]{184}}}
\newcommand{\numfour}{\textcolor{green!60!black}{\ding[1.2]{185}}}
\newcommand{\numfive}{\textcolor{cyan!60!black}{\ding[1.2]{186}}}
\newcommand{\numonec}{\textcolor{red!90!black}{\ding[1.2]{172}}}
\newcommand{\numtwoc}{\textcolor{orange!90!black}{\ding[1.2]{173}}}
\newcommand{\numthreec}{\textcolor{yellow!60!black}{\ding[1.2]{174}}}
\newcommand{\ccol}[2]{\begingroup\setlength{\fboxsep}{1.5pt}\colorbox{#1}{\strut #2}\endgroup}
\def\BibTeX{{\rm B\kern-.05em{\sc i\kern-.025em b}\kern-.08em
    T\kern-.1667em\lower.7ex\hbox{E}\kern-.125emX}}
\begin{document}
\title{The MAMA-MIA Challenge: Advancing Generalizability and Fairness in Breast MRI Tumor Segmentation and Treatment Response Prediction}
\author{
Lidia Garrucho$^{\ast,\dagger}$,
Smriti Joshi $^{\ast}$,
Kaisar Kushibar,
Richard Osuala,
Maciej Bobowicz,
Xavier Bargalló,
Paulius Jaruševičius,
Kai Geissler,
Raphael Schäfer,
Muhammad Alberb,
Tony Xu,
Anne Martel,
Daniel Sleiman,
Navchetan Awasthi,
Hadeel Awwad,
Joan C. Vilanova,
Robert Martí,
Daan Schouten,
Jeong Hoon Lee,
Mirabela Rusu,
Eleonora Poeta,
Luisa Vargas,
Eliana Pastor,
Maria A. Zuluaga,
Jessica Kächele,
Dimitrios Bounias,
Alexandra Ertl,
Katarzyna Gwoździewicz,
Maria-Laura Cosaka,
Pasant M. Abo-Elhoda,
Sara W. Tantawy,
Shorouq S. Sakrana,
Norhan O. Shawky-Abdelfatah,
Amr Muhammad Abdo-Salem,
Androniki Kozana,
Eugen Divjak,
Gordana Ivanac,
Katerina Nikiforaki,
Michail E. Klontzas,
Rosa García-Dosdá,
Meltem Gulsun-Akpinar,
Oğuz Lafcı,
Carlos Martín-Isla,
Oliver Díaz,
Laura Igual,
and Karim Lekadir%
\thanks{$^{\ast}$L. Garrucho and S. Joshi contributed equally to this work.}
\thanks{
L. Garrucho, S. Joshi, K. Kushibar, R. Osuala, C. Martín-Isla, O. Díaz,
L. Igual, and K. Lekadir are with
Barcelona Artificial Intelligence in Medicine Lab (BCN-AIM),
Facultat de Matemàtiques i Informàtica,
Universitat de Barcelona,
Barcelona, Spain.
}
\thanks{
M. Bobowicz and K. Gwoździewicz are with
2nd Department of Radiology,
Medical University of Gdansk,
Gdansk, Poland.
}
\thanks{
X. Bargalló is with
Department of Radiology,
Hospital Clínic of Barcelona,
Barcelona, Spain.
}
\thanks{
P. Jaruševičius is with
Department of Radiology,
Lithuanian University of Health Sciences,
Kaunas, Lithuania.
}
\thanks{
K. Geissler and R. Schäfer are with
Fraunhofer Institute for Digital Medicine MEVIS,
Germany.
}
\thanks{
M. Alberb and T. Xu are with
Department of Medical Biophysics,
University of Toronto,
Canada.
}
\thanks{
A. Martel is with
Sunnybrook Research Institute,
Toronto, Canada.
}
\thanks{
D. Sleiman is with
University of Amsterdam,
Amsterdam, The Netherlands.
}
\thanks{
N. Awasthi is with
University of Amsterdam
and Amsterdam UMC,
Amsterdam, The Netherlands, and with
Indian Institute of Technology, Jodhpur, Rajasthan, India.
}
\thanks{
J. C. Vilanova is with
Department of Radiology, Clínica Girona, Institute of Diagnostic Imaging (IDI), Girona, Spain.
}
\thanks{
H. Awwad and R. Martí are with
Computer Vision and Robotics Institute (ViCOROB),
University of Girona,
Girona, Spain.
}
\thanks{
D. Schouten, J. H. Lee, and M. Rusu are with
Stanford University,
Stanford, CA, USA.
}
\thanks{
E. Poeta and E. Pastor are with
Politecnico di Torino,
Turin, Italy.
}
\thanks{
L. Vargas and M. A. Zuluaga are with
EURECOM,
France.
}
\thanks{
J. Kächele, D. Bounias, and A. Ertl are with
German Cancer Research Center (DKFZ),
Division of Medical Image Computing,
and the Medical Faculty Heidelberg,
Heidelberg University,
Germany.
}
\thanks{
J. Kächele is also with
German Cancer Consortium (DKTK), DKFZ, Core Center Heidelberg, Heidelberg, Germany.
}
\thanks{
M.-L. Cosaka is with
Centro Mamario Instituto Alexander Fleming,
Buenos Aires, Argentina.
}
\thanks{
P. M. Abo-Elhoda, S. W. Tantawy, S. S. Sakrana,
N. O. Shawky-Abdelfatah, and A. M. Abdo-Salem are with
Department of Diagnostic and Interventional Radiology and Molecular Imaging,
Faculty of Medicine,
Ain Shams University,
Cairo, Egypt.
}
\thanks{
A. Kozana is with
Department of Radiology,
University Hospital of Heraklion,
Heraklion, Greece.
}
\thanks{
E. Divjak and G. Ivanac are with
Department of Diagnostic and Interventional Radiology,
University Hospital Dubrava,
and the University of Zagreb School of Medicine,
Zagreb, Croatia.
}
\thanks{
K. Nikiforaki is with
Computational BioMedicine Laboratory,
Institute of Computer Science,
Foundation for Research and Technology--Hellas,
Heraklion, Greece.
}
\thanks{
M. E. Klontzas is with
Department of Radiology,
School of Medicine,
University of Crete,
Heraklion, Greece.
}
\thanks{
R. García-Dosdá is with
Medical Imaging and Radiology,
Universitary and Politechnic Hospital La Fe,
Valencia, Spain.
}
\thanks{
M. Gulsun-Akpinar is with
Department of Radiology,
Hacettepe University Faculty of Medicine,
Ankara, Turkey.
}
\thanks{
O. Lafcı is with
Department of Biomedical Imaging and Image-guided Therapy,
Medical University of Vienna,
Vienna, Austria.
}
\thanks{
K. Lekadir is also with
Institució Catalana de Recerca i Estudis Avançats (ICREA),
Barcelona, Spain.
}
\thanks{
Benchmark team participation:
MIC team (J. Kächele, D. Bounias, A. Ertl);
AIH-MAMA team (E. Poeta, L. Vargas, E. Pastor, M. A. Zuluaga);
pimed-lab team (D. Schouten, J. H. Lee, M. Rusu);
ViCOROB team (H. Awwad, J. C. Vilanova, R. Martí);
AI Strollers team (D. Sleiman, N. Awasthi);
Martel Lab team (M. Alberb, T. Xu, A. Martel);
FME team (K. Geissler, R. Schäfer).
}
\thanks{
This work involved human subjects or animals in its research.
The authors confirm that all human/animal subject research procedures
and protocols were exempt from review board approval.
}
\thanks{$^\dagger$ Corresponding author: Lidia Garrucho (lgarrucho@ub.edu).}
}

\maketitle

\begin{abstract}
Breast cancer is the most frequently diagnosed malignancy among women worldwide and a leading cause of cancer-related mortality. Dynamic contrast-enhanced magnetic resonance imaging plays a central role in tumor characterization and treatment monitoring, particularly in patients receiving neoadjuvant chemotherapy. However, existing artificial intelligence models for breast magnetic resonance imaging are typically developed and evaluated using heterogeneous datasets, study populations, and assessment protocols, making direct comparison difficult and limiting understanding of model robustness across institutions and clinically relevant patient subgroups. The MAMA-MIA Challenge was designed to address these challenges by providing a standardized benchmark for the joint evaluation of primary tumor segmentation and prediction of pathologic complete response using pre-treatment magnetic resonance imaging only. The training cohort comprised 1,506 patients from multiple institutions in the United States, while evaluation was conducted on an external test set of 574 patients from three independent European centers to assess cross-continental and cross-institutional generalization. A unified scoring framework combined predictive performance with subgroup consistency across age, menopausal status, and breast density. Twenty-one international teams participated in the final evaluation phase. Results demonstrate substantial performance variability under a common external evaluation framework and reveal trade-offs between overall accuracy and subgroup fairness. The challenge provides standardized datasets, evaluation protocols, and public resources to promote the development of robust and equitable artificial intelligence systems for breast cancer imaging.
\end{abstract}

\begin{IEEEkeywords}
benchmark, breast cancer, fairness, magnetic resonance imaging, neoadjuvant chemotherapy, pathologic complete response.
\end{IEEEkeywords}

\section{Introduction}
\label{sec:introduction}
Breast cancer is the most frequently diagnosed cancer among women worldwide, with 2.3 million new cases reported in 2022 and persistent disparities in outcomes across regions and populations \cite{bib1,bib2}. Dynamic contrast-enhanced magnetic resonance imaging (DCE-MRI) plays a central role in tumor characterization and treatment monitoring, particularly in the neoadjuvant chemotherapy (NAC) setting \cite{bib3,bib4}. Pathological complete response (pCR) following NAC is a well-established surrogate marker of favorable long-term outcomes, motivating increasing interest in imaging-based pCR prediction from pre-treatment MRI \cite{bib5}.

Artificial intelligence (AI) methods have demonstrated strong potential for automating breast tumor segmentation \cite{bib6, bib7} and predicting treatment response from DCE-MRI \cite{bib8,bib9,bib10,bib11}. Deep learning approaches based on U-Net architectures and their variants have achieved increasingly accurate tumor delineation, supporting downstream quantitative analysis and treatment monitoring \cite{bib12,bib13,bib14,bib6}. However, reported performance varies substantially across datasets, acquisition protocols, and institutions \cite{bib15}, and evaluation practices remain heterogeneous, complicating direct comparison of methods and assessment of real-world robustness \cite{bib16,bib17}.

Similarly, substantial progress has been made in predicting treatment response from pre-treatment breast MRI. Earlier studies primarily relied on radiomics-based approaches, while more recent work has gradually moved to deep learning strategies integrating multimodal data\cite{bib18,bib19,bib20}. Recent work has explored a range of increasingly sophisticated strategies including MRI-based quantification of intratumoral heterogeneity~\cite{bib8}, heterogeneity-informed nomograms integrating imaging biomarkers~\cite{bib9}, and multimodal frameworks combining mammography, ultrasound, MRI for HER2 prediction~\cite{bib10}. Importantly, many of these approaches have moved beyond single-center evaluation by validating their models on independent public cohorts, such as Duke Breast Cancer MRI ~\cite{bib21} and I-SPY2~\cite{bib22}, reflecting an increasing emphasis on assessing model generalizability. This demonstrates the growing maturity of AI-based response prediction and the increasing adoption of external validation practices.

Despite these advances, existing studies are typically developed and evaluated using different patient populations, imaging protocols, outcome definitions, and evaluation methodologies. Consequently, reported results are difficult to compare directly, and it remains unclear how competing approaches perform under a common evaluation framework. Moreover, while demographic and biological factors such as age, menopausal status, breast density, and tumor subtype are increasingly reported and analyzed, subgroup-specific performance assessment remains inconsistent across studies, limiting systematic evaluation of robustness and fairness.

Beyond breast MRI, several community benchmarks have been introduced to promote reproducible and generalizable AI development in medical imaging \cite{bib23,bib24,bib25}. At the same time, growing evidence indicates that AI systems may exhibit performance disparities across demographic and clinical subgroups, motivating the incorporation of fairness-oriented analyses alongside conventional accuracy metrics \cite{bib26,bib27,bib28,bib29}. However, fairness-aware benchmarking remains uncommon in breast imaging, particularly in conjunction with external multi-institutional validation.

To address these challenges, we introduce the \textbf{MAMA-MIA Challenge}, a multi-country benchmark for breast cancer analysis using DCE-MRI. Rather than proposing a single predictive model, the benchmark provides a standardized framework for evaluating and comparing AI methods across institutions and populations. The benchmark jointly evaluates (i) primary tumor segmentation and (ii) prediction of pathological complete response using pre-treatment imaging only. Models are trained on a large multi-institutional United States cohort and evaluated on an external European test set collected from three independent centers in different countries, enabling rigorous assessment of cross-institutional and cross-continental generalization. The evaluation framework further quantifies performance consistency across clinically relevant subgroups, explicitly incorporating fairness alongside predictive accuracy. The benchmark remains active on the CodaBench platform\footnote{\url{https://www.codabench.org/competitions/7425/}}.

\section{Methodology}
\label{sec:methodology}

\subsection{Benchmark Overview}

We define a benchmark framework for advancing computational methods in breast cancer analysis using dynamic contrast-enhanced magnetic resonance imaging (DCE-MRI).
Participants were tasked with addressing two complementary objectives: (i) automatic segmentation of the primary breast tumor (\textit{Task 1}) and (ii) prediction of pathologic complete response (pCR) to neoadjuvant chemotherapy (\textit{Task 2}).

Model development and training relied on large-scale, multi-center breast DCE-MRI data collected across 25 institutions in the United States. Submitted methods were evaluated on private validation and test datasets acquired from three independent European clinical centers in Spain, Poland and Lithuania.  
This design is to enable systematic assessment of cross-institutional generalizability and supports the analysis of performance disparities across clinically relevant demographic subgroups \cite{bib30}.

To this end, Table~\ref{tab1} summarizes the composition and key characteristics of the training, validation, and test cohorts.

\begin{table*}[t]
\centering
\large
\caption{\label{tab1}Clinical and imaging variables across the MAMA-MIA benchmark datasets \cite{bib31}. Four cohorts are used for training (NACT, ISPY1, ISPY2, DUKE) and three for validation/test (GUM, KAU, HCB). Data cells with more than 3 items are split into two rows to maximize readability. Best viewed in color.}
\renewcommand{\arraystretch}{1.5}
\setlength{\tabcolsep}{3pt}

\resizebox{\textwidth}{!}{
\begin{tabular}{|l|rrrrr|rrrr|r|}
\hline
\multicolumn{1}{|c|}{\textbf{MAMA-MIA Dataset}} & \multicolumn{5}{c|}{\textbf{Training}} & \multicolumn{4}{c|}{\textbf{Validation / test}} & \multicolumn{1}{c|}{\textbf{Total}}\\
\cline{1-11}
Collection & NACT & ISPY1 & ISPY2 & DUKE & \multicolumn{1}{c|}{Sum} & GUM & KAU & HCB & \multicolumn{1}{c|}{Sum} & \multicolumn{1}{c|}{Train+Val+Test}\\
\hline
\rowcolor{gray!5}
Country & USA & USA & USA & USA &  & Poland & Lithuania & Spain &  &  \\
Year & 2002–06 & 2010–16 & 2000–14 & 1995–02 &  & 2010–20 & 2010–20 & 2010–20 &  &  \\
\hline
\multicolumn{11}{|l|}{\textbf{Clinical variables}}\\
\hline
\rowcolor{gray!5}
\shortstack[l]{\textbf{pCR}\\\ccol{colA}{Yes}/\ccol{colB}{No}/\ccol{colC}{N.A.}} &
\ccol{colA}{11}/\ccol{colB}{53}/\ccol{colC}{0} & 
\ccol{colA}{49}/\ccol{colB}{118}/\ccol{colC}{4} & 
\ccol{colA}{316}/\ccol{colB}{664}/\ccol{colC}{0} & 
\ccol{colA}{64}/\ccol{colB}{216}/\ccol{colC}{11} & 
\ccol{colA}{440}/\ccol{colB}{1051}/\ccol{colC}{15} & 
\ccol{colA}{11}/\ccol{colB}{19}/\ccol{colC}{--} & 
\ccol{colA}{71}/\ccol{colB}{161}/\ccol{colC}{--} & 
\ccol{colA}{78}/\ccol{colB}{234}/\ccol{colC}{--} & 
\ccol{colA}{160}/\ccol{colB}{414}/\ccol{colC}{--} & 
\ccol{colA}{600}/\ccol{colB}{1465}/\ccol{colC}{15}\\

\shortstack[l]{\textbf{Tumor subtype} \\ \ccol{colA}{Lum} / \ccol{colB}{TN} / \ccol{colC}{HER2enr} \\ \ccol{colD}{HER2pure} / \ccol{colE}{N.A.}} &
\shortstack[r]{\ccol{colA}{21} / \ccol{colB}{11} / \ccol{colC}{8} \\ \ccol{colD}{6} / \ccol{colE}{18}} & 
\shortstack[r]{\ccol{colA}{67} / \ccol{colB}{45} / \ccol{colC}{25} \\ \ccol{colD}{29} / \ccol{colE}{5}} & 
\shortstack[r]{\ccol{colA}{536} / \ccol{colB}{358} / \ccol{colC}{86} \\ \ccol{colD}{0} / \ccol{colE}{0}} & 
\shortstack[r]{\ccol{colA}{123} / \ccol{colB}{85} / \ccol{colC}{50} \\ \ccol{colD}{30} / \ccol{colE}{3}} & 
\shortstack[r]{\ccol{colA}{747} / \ccol{colB}{499} / \ccol{colC}{169} \\ \ccol{colD}{65} / \ccol{colE}{26}} & 
\shortstack[r]{\ccol{colA}{11} / \ccol{colB}{9} / \ccol{colC}{3} \\ \ccol{colD}{7} / \ccol{colE}{--}} & 
\shortstack[r]{\ccol{colA}{95} / \ccol{colB}{71} / \ccol{colC}{39} \\ \ccol{colD}{27} / \ccol{colE}{--}} & 
\shortstack[r]{\ccol{colA}{209} / \ccol{colB}{43} / \ccol{colC}{49} \\ \ccol{colD}{11} / \ccol{colE}{--}} & 
\shortstack[r]{\ccol{colA}{315} / \ccol{colB}{123} / \ccol{colC}{91} \\ \ccol{colD}{45} / \ccol{colE}{--}} & 
\shortstack[r]{\ccol{colA}{1062} / \ccol{colB}{622} / \ccol{colC}{260} \\ \ccol{colD}{110} / \ccol{colE}{26}}\\

\rowcolor{gray!5}
\shortstack[l]{\textbf{Age group} \\ \ccol{colA}{$<$40}/\ccol{colB}{40–49}/\ccol{colC}{50–59} \\ \ccol{colD}{60–69}/\ccol{colE}{$>$70}/\ccol{colF}{N.A.}} &
\shortstack[r]{\ccol{colA}{13}/\ccol{colB}{25}/\ccol{colC}{18}\\\ccol{colD}{6}/\ccol{colE}{2}/\ccol{colF}{0}} & 
\shortstack[r]{\ccol{colA}{35}/\ccol{colB}{62}/\ccol{colC}{56}\\\ccol{colD}{18}/\ccol{colE}{0}/\ccol{colF}{0}} & 
\shortstack[r]{\ccol{colA}{208}/\ccol{colB}{300}/\ccol{colC}{317}\\\ccol{colD}{134}/\ccol{colE}{18}/\ccol{colF}{3}} & 
\shortstack[r]{\ccol{colA}{61}/\ccol{colB}{104}/\ccol{colC}{74}\\\ccol{colD}{41}/\ccol{colE}{11}/\ccol{colF}{0}} & 
\shortstack[r]{\ccol{colA}{317}/\ccol{colB}{491}/\ccol{colC}{465}\\\ccol{colD}{199}/\ccol{colE}{31}/\ccol{colF}{3}} & 
\shortstack[r]{\ccol{colA}{7}/\ccol{colB}{9}/\ccol{colC}{6}\\\ccol{colD}{7}/\ccol{colE}{1}/\ccol{colF}{--}} & 
\shortstack[r]{\ccol{colA}{21}/\ccol{colB}{51}/\ccol{colC}{54}\\\ccol{colD}{61}/\ccol{colE}{45}/\ccol{colF}{--}} & 
\shortstack[r]{\ccol{colA}{22}/\ccol{colB}{57}/\ccol{colC}{86}\\\ccol{colD}{87}/\ccol{colE}{60}/\ccol{colF}{--}} & 
\shortstack[r]{\ccol{colA}{50}/\ccol{colB}{117}/\ccol{colC}{146}\\\ccol{colD}{155}/\ccol{colE}{106}/\ccol{colF}{--}} & 
\shortstack[r]{\ccol{colA}{367}/\ccol{colB}{608}/\ccol{colC}{611}\\\ccol{colD}{354}/\ccol{colE}{137}/\ccol{colF}{3}}\\

\shortstack[l]{\textbf{Menopausal status}\\\ccol{colA}{Pre}/\ccol{colB}{Post}/\ccol{colC}{N.A.}} &
\ccol{colA}{0}/\ccol{colB}{0}/\ccol{colC}{64} & 
\ccol{colA}{0}/\ccol{colB}{0}/\ccol{colC}{171} & 
\ccol{colA}{512}/\ccol{colB}{296}/\ccol{colC}{172} & 
\ccol{colA}{166}/\ccol{colB}{124}/\ccol{colC}{1} & 
\ccol{colA}{678}/\ccol{colB}{420}/\ccol{colC}{408} & 
\ccol{colA}{17}/\ccol{colB}{13}/\ccol{colC}{--} & 
\ccol{colA}{73}/\ccol{colB}{159}/\ccol{colC}{--} & 
\ccol{colA}{109}/\ccol{colB}{203}/\ccol{colC}{--} & 
\ccol{colA}{199}/\ccol{colB}{375}/\ccol{colC}{--} & 
\ccol{colA}{877}/\ccol{colB}{795}/\ccol{colC}{408}\\
\hline
\multicolumn{11}{|l|}{\textbf{Imaging variables}}\\
\hline
\rowcolor{gray!5}
\shortstack[l]{\textbf{Acq. Plane}\\\ccol{colA}{Axial}/\ccol{colB}{Sagittal}} &
\ccol{colA}{0}/\ccol{colB}{64} & 
\ccol{colA}{0}/\ccol{colB}{171} & 
\ccol{colA}{980}/\ccol{colB}{0} & 
\ccol{colA}{291}/\ccol{colB}{0} & 
\ccol{colA}{1271}/\ccol{colB}{235} & 
\ccol{colA}{30}/\ccol{colB}{0} & 
\ccol{colA}{225}/\ccol{colB}{7} & 
\ccol{colA}{312}/\ccol{colB}{0} & 
\ccol{colA}{567}/\ccol{colB}{7} & 
\ccol{colA}{1838}/\ccol{colB}{242}\\

\shortstack[l]{\textbf{Field strength (T)}\\\ccol{colA}{1.5}/\ccol{colB}{3.0}} &
\ccol{colA}{64}/\ccol{colB}{0} & 
\ccol{colA}{171}/\ccol{colB}{0} & 
\ccol{colA}{715}/\ccol{colB}{265} & 
\ccol{colA}{136}/\ccol{colB}{155} & 
\ccol{colA}{1086}/\ccol{colB}{420} & 
\ccol{colA}{30}/\ccol{colB}{0} & 
\ccol{colA}{27}/\ccol{colB}{205} & 
\ccol{colA}{283}/\ccol{colB}{29} & 
\ccol{colA}{340}/\ccol{colB}{234} & 
\ccol{colA}{1426}/\ccol{colB}{654}\\

\rowcolor{gray!5}
\shortstack[l]{\textbf{Fat suppressed}\\\ccol{colA}{Yes}/\ccol{colB}{No}} &
\ccol{colA}{64}/\ccol{colB}{0} & 
\ccol{colA}{170}/\ccol{colB}{1} & 
\ccol{colA}{976}/\ccol{colB}{4} & 
\ccol{colA}{290}/\ccol{colB}{1} & 
\ccol{colA}{1500}/\ccol{colB}{6} & 
\ccol{colA}{30}/\ccol{colB}{0} & 
\ccol{colA}{232}/\ccol{colB}{0} & 
\ccol{colA}{312}/\ccol{colB}{0} & 
\ccol{colA}{574}/\ccol{colB}{0} & 
\ccol{colA}{2074}/\ccol{colB}{6}\\

\shortstack[l]{\textbf{Manufacturer}\\\ccol{colA}{Siemens}/\ccol{colB}{GE}\\\ccol{colC}{Philips}} &
\ccol{colA}{0}/\ccol{colB}{64}/\ccol{colC}{0} & 
\ccol{colA}{44}/\ccol{colB}{115}/\ccol{colC}{12} & 
\ccol{colA}{252}/\ccol{colB}{611}/\ccol{colC}{117} & 
\ccol{colA}{115}/\ccol{colB}{176}/\ccol{colC}{0} & 
\ccol{colA}{411}/\ccol{colB}{966}/\ccol{colC}{129} & 
\ccol{colA}{30}/\ccol{colB}{0}/\ccol{colC}{0} & 
\ccol{colA}{27}/\ccol{colB}{0}/\ccol{colC}{205} & 
\ccol{colA}{96}/\ccol{colB}{216}/\ccol{colC}{0} & 
\ccol{colA}{153}/\ccol{colB}{216}/\ccol{colC}{205} & 
\ccol{colA}{564}/\ccol{colB}{1182}/\ccol{colC}{334}\\

\rowcolor{gray!5}
\shortstack[l]{\textbf{Slice thickness (mm)}\\\ccol{colA}{$<$1}/\ccol{colB}{1–1.49}/\ccol{colC}{1.5–1.99}\\\ccol{colD}{2–2.49}/\ccol{colE}{$\geq$2.5}} &
\shortstack[r]{\ccol{colA}{0}/\ccol{colB}{0}/\ccol{colC}{0}\\\ccol{colD}{64}/\ccol{colE}{0}} & 
\shortstack[r]{\ccol{colA}{0}/\ccol{colB}{0}/\ccol{colC}{5}\\\ccol{colD}{83}/\ccol{colE}{83}} & 
\shortstack[r]{\ccol{colA}{11}/\ccol{colB}{163}/\ccol{colC}{8}\\\ccol{colD}{695}/\ccol{colE}{103}} & 
\shortstack[r]{\ccol{colA}{0}/\ccol{colB}{283}/\ccol{colC}{4}\\\ccol{colD}{3}/\ccol{colE}{1}} & 
\shortstack[r]{\ccol{colA}{11}/\ccol{colB}{446}/\ccol{colC}{17}\\\ccol{colD}{845}/\ccol{colE}{187}} & 
\shortstack[r]{\ccol{colA}{0}/\ccol{colB}{30}/\ccol{colC}{0}\\\ccol{colD}{0}/\ccol{colE}{0}} & 
\shortstack[r]{\ccol{colA}{27}/\ccol{colB}{0}/\ccol{colC}{196}\\\ccol{colD}{9}/\ccol{colE}{0}} & 
\shortstack[r]{\ccol{colA}{0}/\ccol{colB}{2}/\ccol{colC}{36}\\\ccol{colD}{272}/\ccol{colE}{2}} & 
\shortstack[r]{\ccol{colA}{27}/\ccol{colB}{32}/\ccol{colC}{232}\\\ccol{colD}{281}/\ccol{colE}{2}} & 
\shortstack[r]{\ccol{colA}{38}/\ccol{colB}{478}/\ccol{colC}{249}\\\ccol{colD}{1126}/\ccol{colE}{189}}\\

\shortstack[l]{\textbf{Pixel spacing (mm)}\\\ccol{colA}{$<$0.5}/\ccol{colB}{0.5–0.99}\\\ccol{colC}{1–1.49}} &
\ccol{colA}{2}/\ccol{colB}{62}/\ccol{colC}{0} & 
\ccol{colA}{15}/\ccol{colB}{150}/\ccol{colC}{6} & 
\ccol{colA}{2}/\ccol{colB}{939}/\ccol{colC}{39} & 
\ccol{colA}{0}/\ccol{colB}{274}/\ccol{colC}{17} & 
\ccol{colA}{19}/\ccol{colB}{1425}/\ccol{colC}{62} & 
\ccol{colA}{0}/\ccol{colB}{1}/\ccol{colC}{29} & 
\ccol{colA}{8}/\ccol{colB}{224}/\ccol{colC}{0} & 
\ccol{colA}{0}/\ccol{colB}{306}/\ccol{colC}{6} & 
\ccol{colA}{8}/\ccol{colB}{531}/\ccol{colC}{35} & 
\ccol{colA}{27}/\ccol{colB}{1956}/\ccol{colC}{97}\\

\rowcolor{gray!5}
\shortstack[l]{\textbf{Echo time (ms)}\\\ccol{colA}{1–1.99}/\ccol{colB}{2–2.99}\\\ccol{colC}{$\geq$3}} &
\ccol{colA}{0}/\ccol{colB}{0}/\ccol{colC}{64} & 
\ccol{colA}{1}/\ccol{colB}{1}/\ccol{colC}{169} & 
\ccol{colA}{244}/\ccol{colB}{338}/\ccol{colC}{398} & 
\ccol{colA}{115}/\ccol{colB}{176}/\ccol{colC}{0} & 
\ccol{colA}{360}/\ccol{colB}{515}/\ccol{colC}{631} & 
\ccol{colA}{30}/\ccol{colB}{0}/\ccol{colC}{0} & 
\ccol{colA}{20}/\ccol{colB}{205}/\ccol{colC}{7} & 
\ccol{colA}{98}/\ccol{colB}{208}/\ccol{colC}{6} & 
\ccol{colA}{148}/\ccol{colB}{413}/\ccol{colC}{13} & 
\ccol{colA}{508}/\ccol{colB}{928}/\ccol{colC}{644}\\

\shortstack[l]{\textbf{Repetition time (ms)}\\\ccol{colA}{3–3.99}/\ccol{colB}{4–4.99}\\\ccol{colC}{5–5.99}/\ccol{colD}{$\geq$6}} & 
\shortstack[r]{\ccol{colA}{0}/\ccol{colB}{0}\\\ccol{colC}{0}/\ccol{colD}{64}} & 
\shortstack[r]{\ccol{colA}{0}/\ccol{colB}{0}\\\ccol{colC}{1}/\ccol{colD}{170}} & 
\shortstack[r]{\ccol{colA}{1}/\ccol{colB}{226}\\\ccol{colC}{218}/\ccol{colD}{535}} & 
\shortstack[r]{\ccol{colA}{32}/\ccol{colB}{116}\\\ccol{colC}{116}/\ccol{colD}{27}} & 
\shortstack[r]{\ccol{colA}{33}/\ccol{colB}{342}\\\ccol{colC}{335}/\ccol{colD}{796}} & 
\shortstack[r]{\ccol{colA}{30}/\ccol{colB}{0}\\\ccol{colC}{0}/\ccol{colD}{0}} & 
\shortstack[r]{\ccol{colA}{0}/\ccol{colB}{223}\\\ccol{colC}{2}/\ccol{colD}{7}} & 
\shortstack[r]{\ccol{colA}{0}/\ccol{colB}{269}\\\ccol{colC}{34}/\ccol{colD}{9}} & 
\shortstack[r]{\ccol{colA}{30}/\ccol{colB}{492}\\\ccol{colC}{36}/\ccol{colD}{16}} & 
\shortstack[r]{\ccol{colA}{63}/\ccol{colB}{834}\\\ccol{colC}{371}/\ccol{colD}{812}}\\

\rowcolor{gray!5}
\shortstack[l]{\textbf{Image size {[}x, y, z{]}}
\\ Median} 

 & [256,256,60] & [256,256,60] & [512,512,80] & [512,512,170] & [512,512,80] & [352,352,160] & [528,528,187] & [512,512,82] & [512,512,112] & [512, 512, 80]\\
\rowcolor{gray!5}
Mean & [264,264,59] & [278,278,63] & [481,481,105] & [479,479,169] & [449,449,111] & [352,352,166] & [525,525,190] & [485,485,88] & [494,494,134] & [461,461,117]\\
\rowcolor{gray!5}
Std & [44,44,2] & [72,72,28] & [69,69,41] & [49,49,22] & [99,99,48] & [0,2,10] & [71,71,20] & [40,40,24] & [66,66,54] & [94,94,51]\\
\hline
\textbf{Subjects} & \textbf{64} & \textbf{171} & \textbf{980} & \textbf{291} & \textbf{1506} & \textbf{30} & \textbf{232} & \textbf{312} & \textbf{574} & \textbf{2080}\\
\hline
\end{tabular}
}
\end{table*}

\subsubsection{Training Dataset}

The training cohort was based on the publicly available MAMA-MIA dataset~\cite{bib31}, which was respectively built for this purpose and encompassess a multi-center resource comprising 1,506 patients with pre-treatment T1-weighted DCE breast MRI examinations.  
As detailed in ~\cite{bib31}, the MAMA-MIA dataset aggregates and harmonizes data from multiple TCIA \cite{bib32} collections, including ISPY-1~\cite{bib33}, ISPY-2~\cite{bib22}, NACT~\cite{bib34}, and DUKE~\cite{bib21}.

Expert annotations are provided for the training data, including voxel-wise primary tumor segmentation masks and bounding boxes indicating the affected breast. Relevant clinical metadata, such as age, menopausal status, and tumor subtype, are available for training to support subgroup analysis and bias-aware model development. This metadata is withheld at test time to ensure that evaluation relies exclusively on imaging data.

The training dataset further exhibits substantial heterogeneity in acquisition parameters. Data were acquired from multiple vendors (GE, Siemens, Philips) and field strengths (1.5T and 3T), reflecting substantial acquisition heterogeneity.

In addition to the MAMA-MIA dataset, participants were permitted to use any publicly available datasets released under CC-BY-NC or more permissive licenses for model training.

\subsubsection{Validation and test Datasets}

The validation and test datasets are private and consist of 574 fat-suppressed DCE-MRI examinations collected from three European clinical centers: Medical University of Gdansk (GUM, Poland; 30 cases), Kauno Klinikos (KAU, Lithuania; 232 cases), and Hospital Clínic Barcelona (HCB, Spain; 312 cases).

Imaging was predominantly acquired in the axial plane (98.8\%), with a small proportion in the sagittal plane (1.2\%). Scans were obtained using 1.5T (59.2\%) and 3T (40.8\%) systems across multiple vendors, including Siemens (26.7\%), GE (37.6\%), and Philips (35.7\%).

The dataset was split into a validation set of 58 cases and a held-out test set of 516 cases. Splitting was stratified by center, age, breast density, and menopausal status to ensure well-balanced representation in test and validation across these clinically relevant subgroups.

To assist teams in designing robust pipelines for both tasks, full cohort-level summaries of both the training and test datasets were made publicly available from the outset. For the classification task, the prevalence of target labels was kept consistent across the training and test splits, allowing participants to focus on model generalization and target-specific challenges (such as data imbalance) without experiencing significant label distribution shifts.

\subsection{Annotation Protocol}

\subsubsection{Training Dataset}

For the training data, the primary tumor was annotated on the first post-contrast DCE-MRI volume through a multi-stage workflow. First, an initial automatic segmentation was generated using a baseline nnU-Net model~\cite{bib13}. These automatic segmentations were then clinically reviewed by two expert clinicians, who assigned a quality label (Good, Acceptable, Poor, or Missed). Based on these labels, the cases were distributed among a panel of 16 expert clinicians from 10 institutions across Europe, Asia, and Africa, who manually corrected the segmentations as needed by adding or removing regions to obtain the final annotations. Following collection of the expert-corrected annotations, the authors performed a technical quality-control pass to verify data integrity (e.g., file format consistency and the absence of corrupted or invalid segmentations), without modifying the experts' delineations. No additional clinical approval step was performed after the expert corrections. Further details regarding the annotation guidelines and quality-control workflow are provided in the MAMA-MIA dataset publication~\cite{bib31}.

\subsubsection{Validation and Test Datasets}

Primary tumor segmentations for the validation and test datasets were produced by four expert clinicians from the corresponding clinical centers (GUM, KAU, and HCB). To reduce inter-annotator variability, a harmonized annotation protocol was established, in which pathological boundaries were standardized through a pre-annotation calibration workshop~\cite{bib35}. Segmentation was performed using a cloud-based collaborative annotation platform~\footnote{https://www.cmrad.com/}, and tumor volumes were initialized with a shared 3D Smart Paint tool~\cite{bib36}. This tool provides interactive, user-guided tumor segmentation: clinicians first define a coarse delineation of the lesion on the slice with the highest conspicuity, after which the algorithm generates an initial 3D tumor region consistent with local image context across adjacent slices. These initial predictions are then iteratively refined through user interaction, where clinicians add or remove regions via a painting-based interface until consensus is reached.

\subsection{Selection of Fairness Variables}

To assess potential subgroup disparities in model performance, fairness analyses were conducted across age, menopausal status, and breast density groups. These variables were selected following recent recommendations for fairness evaluation in medical AI, which emphasize subgroup-based assessment across clinically relevant patient populations \cite{bib37}. Age, menopausal status, and breast density are routinely reported patient-level characteristics in breast imaging and are associated with differences in breast cancer risk, hormonal environment and breast tissue composition \cite{bib38,bib39,bib40, bib41, bib42, bib43}. Importantly, these variables were included as stratification factors for fairness assessment rather than as assumed causal determinants of model performance trained on Breast MRI.

We note that additional variables, such as molecular subtype (e.g., HR/HER2-defined subtypes), may also influence imaging appearance and treatment response. Accordingly, reporting performance stratified by subtype is common practice in the literature. However, unlike age, menopausal status, and breast density, tumor subtype is a disease-specific biological characteristic that is strongly associated with the target outcome, particularly in the pCR prediction task \cite{bib44,bib45}. Therefore, although we report subtype-specific performance to facilitate comparison with prior studies, molecular subtype was not included in the computation of the fairness score.

\subsection{Data Preprocessing}

All training, validation, and test cases underwent standardized preprocessing to ensure consistent data handling across sites. Raw DICOM files were converted to NIfTI format using the \textit{pycad} library~\cite{bib46}. Dataset harmonization included quality control to remove scans with artifacts or incomplete sequences, extraction of relevant metadata, and enforcement of a uniform naming convention and folder structure \cite{bib31}.

Image orientation was standardized to preserve anatomical consistency: sagittal acquisitions were reoriented to the posterior–superior–right (PSR) coordinate system, and axial acquisitions to the left–anterior–superior (LAS) system.
No additional preprocessing steps, such as intensity normalization, bias field correction, or voxel resampling, were enforced at the benchmark level. This design choice avoids introducing assumptions that may favor specific modeling strategies and allows participants to apply task-specific preprocessing tailored to their methods.

\subsection{Benchmark Setup}

All submissions were evaluated automatically using a standardized Python-based evaluation pipeline executed within a containerized environment on the CodaBench platform \cite{bib47}. Participants were required to submit a single \texttt{.zip} archive containing source code and any required model weights, following a predefined submission template to ensure reproducibility and fair comparison.

The MAMA-MIA benchmark challenge was organized in multiple phases. An initial \textit{Sanity-Check} phase using training data was used to validate submission formatting and code execution. This was followed by a validation phase in external data, during which teams were limited to one submission per day, with a maximum of 15 submissions per team, to mitigate overfitting.  
The benchmark concluded with a final test phase, in which performance was assessed on a held-out test set and only a single submission per team was permitted. To support participants throughout the development process, dedicated communication channels were maintained by the organizers. All evaluations were executed locally on a Linux system equipped with an NVIDIA GeForce RTX 3090 GPU with 24~GB of memory. To ensure computational feasibility during large-scale evaluation and to encourage clinically practical solutions, submissions were required to complete inference within a maximum execution time of 5 minutes per image. Submissions exceeding this limit were automatically terminated and considered invalid.

The benchmark challenge was initially introduced in conjunction with the MICCAI 2025 conference \cite{bib48}, where it was hosted as part of the 2nd Deep-Breath Workshop \cite{bib49}. The challenge website, documentation, and resources are publicly available at \url{https://www.ub.edu/mama-mia}.

\subsection{Evaluation}

Participants were encouraged to participate in both tasks: primary tumor segmentation (\textit{Task 1}) and pathologic complete response (pCR) prediction (\textit{Task 2}).  
Although the tasks differ in nature, both were evaluated using a unified scoring framework that jointly accounts for predictive performance and algorithmic fairness across clinically relevant demographic subgroups.

\subsubsection{Unified Scoring Framework}

For each task, submissions were ranked using a combined score that integrates task-specific predictive performance with fairness:
\begin{equation}
\label{eq:overall_score}
\mathcal{S}
=
(1 - \lambda)\,\mathcal{S}_{p}
+
\lambda\,\mathcal{S}_{f},
\end{equation}
where $\mathcal{S}_{p}$ denotes the performance score, $\mathcal{S}_{f}$ the fairness score, and $\lambda \in [0,1]$ controls the trade-off between the two.  
In this benchmark, we set $\lambda = 0.5$, assigning equal importance to predictive accuracy and fairness.

Fairness was evaluated across a shared set of demographic variables:
\begin{equation}
V = \{ \text{age},\ \text{menopausal status},\ \text{breast density} \},
\end{equation}

For each fairness variable $v \in V$ , let $G_{v}$ denote the set of subgroups induced by $v$.

\begin{subequations}
\label{eq:subgroups}
\begin{align}
\text{$\mathcal{G}_{age}$}
&= \{ <40,\ 41\!-\!50,\ 51\!-\!60,\ 61\!-\!70,\ >71\}, \\
\text{$\mathcal{G}_{menopausal}$}
&= \{ \text{pre},\ \text{post} \}, \\
\text{$\mathcal{G}_{density}$}
&= \{ \text{A},\ \text{B},\ \text{C},\ \text{D} \}.
\end{align}
\end{subequations}

This formulation incentivizes models to achieve consistent performance across subgroups of fairness variables rather than optimizing for average accuracy alone. Breast density was categorized according to BI-RADS breast composition categories (A–D: fatty, scattered fibroglandular density, heterogeneously dense, and extremely dense breasts)\cite{bib50}. 

\subsubsection{Task 1: Primary Tumor Segmentation}

Segmentation performance was assessed using two complementary metrics: the Dice Similarity Coefficient (DSC) and the normalized Hausdorff Distance (NormHD).  
The performance score is defined as:
\begin{subequations}
\label{eq:seg_perf}
\begin{align}
\mathcal{S}_{p}^{\text{seg}}
&= \frac{1}{2}\left( \text{DSC} + (1 - \text{NormHD}) \right), \\
\text{DSC}
&= \frac{1}{N} \sum_{i=1}^{N}
\frac{2 |P_i \cap Q_i|}{|P_i| + |Q_i|}, \\
\text{NormHD}
&= \frac{1}{N} \sum_{i=1}^{N}
\frac{HD(P_i, Q_i)}{HD_{\max}},
\end{align}
\end{subequations}
where $N$ are the total number of evaluated cases, $P_i$ and $Q_i$ denote the predicted and ground-truth segmentations for case $i$, respectively, and $HD_{\max}=150$~mm.

Additionally, for a given performance metric $M$, we denote by $\overline{M}_{v,g}$ the average value of $M$ computed over all samples belonging to subgroup $g \in \mathcal{G}_v$, defined in Equation \ref{eq:subgroups}. The disparity $D_{v}$  associated with fairness variable $v \in V$ is then computed as,

\begin{subequations}
\label{eq:seg_fair}
\begin{align}
D_{v} 
&= \frac{1}{2}
\sum_{M \in \{\mathrm{DSC},\ \mathrm{NormHD}\}}
\left(
\max_{g \in \mathcal{G}_v} \overline{M}_{v,g}
-
\min_{g \in \mathcal{G}_v} \overline{M}_{v,g}
\right) \\
\intertext{The fairness score $\mathcal{S}_{f}^{\text{seg}}$ for segmentation over all fairness variables is computed as:}
\mathcal{S}_{f}^{\text{seg}}
&= \frac{1}{|V|}
\sum_{v \in V} (1 - D_v)
\end{align}
\end{subequations}

Empty segmentations were handled by assigning a DSC of 0 and a Hausdorff Distance of $HD_{\max}$, accounting for incorrect or missing predictions in bilateral breast images.  
The value of $HD_{\max}$ was derived from the maximum distance observed in the automatic segmentation baseline.

\subsubsection{Task 2: Pathologic Complete Response Prediction}

Classification performance was evaluated using Balanced Accuracy (BA), defined as:
\begin{subequations}
\label{eq:cls_perf}
\begin{align}
\text{TPR}
&= \frac{1}{N_+} \sum_{i \in \mathcal{P}} \mathbf{1}(y_i = \hat{y}_i), \\
\text{TNR}
&= \frac{1}{N_-} \sum_{i \in \mathcal{N}} \mathbf{1}(y_i = \hat{y}_i), \\
\mathcal{S}_{p}^{\text{cls}}
&= \frac{\text{TPR} + \text{TNR}}{2},
\end{align}
\end{subequations}
where $\mathcal{P}$ and $\mathcal{N}$ denote the sets of positive and negative samples, respectively.
The classification fairness score $\mathcal{S}_{f}^{\text{cls}}$ was defined using an Equalized Odds–based criterion. For each variable $v$:
\begin{subequations}
\label{eq:cls_fair}
\begin{align}
\mathcal{D}^{\mathrm{TPR}}_v
&= \max_{g \in \mathcal{G}_v} \mathrm{TPR}_{v,g}
   - \min_{g \in \mathcal{G}_v} \mathrm{TPR}_{v,g}, \\
\mathcal{D}^{\mathrm{FPR}}_v
&= \max_{g \in \mathcal{G}_v} \mathrm{FPR}_{v,g}
   - \min_{g \in \mathcal{G}_v} \mathrm{FPR}_{v,g}, \\
\mathcal{D}^{\mathrm{EQ}}_v
&= \mathcal{D}^{\mathrm{TPR}}_v
   + \mathcal{D}^{\mathrm{FPR}}_v, \\
\mathcal{S}_{f}^{\text{cls}}
&= 1 - \frac{1}{|V|} \sum_{v \in V} \mathcal{D}^{EQ}_v .
\end{align}
\end{subequations}

This formulation rewards classifiers that maintain consistent error rates across demographic subgroups.  Submissions producing constant predictions (e.g., all zeros or all ones) were considered invalid and excluded from evaluation.

Although not part of the official challenge evaluation framework, we also computed Recall (sensitivity) and the Area Under the Receiver Operating Characteristic Curve (AUC) to compare the different classification methods in Section \ref{sec:results}. Recall is defined as $\text{Recall} = \frac{\text{TP}}{\text{TP} + \text{FN}}$ (where TP are true positives and FN are false negatives). The AUC is mathematically equivalent to the probability that the model ranks a randomly chosen positive instance higher than a randomly chosen negative one, denoted as $\text{AUC} = P(f(x^+) > f(x^-))$, where $f(\cdot)$ is the classifier's output score.

\subsection{Baselines}

For \textit{Task 1} (primary tumor segmentation), the baseline model was a 3D nnUNet trained on the 1,506 DCE-MRI volumes with expert annotations from the MAMA-MIA dataset\cite{bib31}. Given nnUNet's widespread adoption in medical image segmentation benchmarks, this baseline provides a strong reference point while encouraging participants to explore novel architectures and learning strategies. Baseline model weights are publicly available in Synapse \footnote{\url{https://www.synapse.org/Synapse:syn60868042/wiki/628716}}.

For \textit{Task 2} (pCR prediction), the baseline was a random classifier. Specifically, for each test case the classifier draws a prediction score uniformly at random from $[0, 1]$ using \texttt{numpy.random.rand()} function and assigns the binary pCR label by thresholding this score at $0.5$, without using any imaging, segmentation-derived,
or clinical information. The prevalence of pCR was comparable across splits (training: 440/1{,}491 labeled cases, $\approx$29.5\%; validation/test: 160/574, $\approx$27.9\%; Table~\ref{tab1}). This baseline corresponds to the \texttt{predict\_classification} method of the official sample submission provided to all participants, publicly available in the competition CodaBench repository. This model establishes a lower bound on performance and allows participants to quantify the added value of their imaging-driven predictive models.

\subsection{Participating Methods}
\label{sec:participating_methods}

 This section provides a structured overview of the participating methods for each task, with the goal of characterizing the methodological space explored by the community rather than exhaustively describing individual implementations. For each task, we summarize common design choices and methodological trends, followed by concise technical descriptions of the top-ranked approaches. Detailed architectural, preprocessing, and training configurations are reported in Tables~\ref{tab2} and~\ref{tab3}.

\begin{table*}[h]
\caption{Technical summary of top-performing primary tumor segmentation methods (Task 1). Abbreviations used in the table- nnUNet pre: default nnUNet \cite{bib13} preprocessing strategy;  RAS: Right-Anterior-Superior orientation; p0/p1/p2: pre-contrast, first post-contrast and second post-contrast phases; px: last available post-contrast phase; s1: subtraction of first post-contrast phase with pre-contrast phase; bbox: Bounding Box; CE: Cross Entropy; CV: Cross Validation; n/a: not applicable/available.}
\centering
\setlength{\tabcolsep}{5pt}
\definecolor{rowgray}{gray}{0.93}
\rowcolors{3}{rowgray}{white}
\begin{threeparttable}
\begin{tabular}{lccccc}
\toprule
 & \textbf{\numone  \text{ MIC}} & \textbf{ \numtwo  \text{ FME}} & \textbf{\numthree  \text{ ViCOROB}} & \textbf{\numfour  \text{ Martel Lab}} & \textbf{\numfive  \text{ AIH-MAMA}} \\
\midrule
\textbf{Preprocessing Strategy} &&&&&\\
resample &  \begin{tabular}[c]{@{}c@{}}third-order spline \\ $1 \times 1 \times 1$ \end{tabular}  & \begin{tabular}[c]{@{}c@{}} nnUNet pre \\ $0.7 \times 0.7 \times 2.0$ \end{tabular}  & \begin{tabular}[c]{@{}c@{}} b-spline \\ $1 \times 1 \times 1$ \end{tabular} & \begin{tabular}[c]{@{}c@{}} trilinear \\ $1 \times 1 \times 1$ \end{tabular}   & \begin{tabular}[c]{@{}c@{}} bilinear \\ $1 \times 1 \times 1$ \end{tabular} \\
crop or pad & \xmark & \xmark & breast bbox & breast bbox &   \begin{tabular}[c]{@{}c@{}}
non-zero voxels\\
$320 \times 320 \times 128$
\end{tabular} \\
reorient & \xmark & \xmark & \xmark & RAS & RAS \\
intensity clipping & \xmark & \xmark & \begin{tabular}[c]{@{}c@{}} percentile \\$[0.1, 99.9]$ \end{tabular} & \begin{tabular}[c]{@{}c@{}} percentile \\$[0.05, 99.95]$ \end{tabular}  & \xmark \\
normalize
& z-score
& z-score
& z-score
& min-max
&  \begin{tabular}[c]{@{}c@{}} z-score \\ (non-zero pixels only)\end{tabular} \\
\midrule
\textbf{Training Strategy} &&&&& \\
DCE phases
& p0, p1, p2
& s1
& p0, p1, p2
& p0, p1, px
& p0, p1, p2 \\
network
& \begin{tabular}[c]{@{}c@{}}3D nnUNet  \\ (ResEncL) \cite{bib13}\end{tabular}
& \begin{tabular}[c]{@{}c@{}}3D nnUNet \\ (ResEncL) \cite{bib13}\end{tabular} 
& \begin{tabular}[c]{@{}c@{}}3D nnUNet \\ (ResEncL) \cite{bib13}\end{tabular} 
& 3D ViT \cite{bib58}
& 3D UNet \\
\# parameters (M)\tnote{a}
& 383.5
& 2637.35
& 102.35
& 597.58
& 4.8 \\
pretrained & \cmark & \xmark & \xmark & \cmark & \xmark \\
optimizer & SGD \cite{bib51} & SGD\cite{bib51} & SGD\cite{bib51} & AdamW \cite{bib52} & AdamW \cite{bib52} \\
batch size & 3 & 2 & 3 & 8 & 1 \\
loss
& Soft Dice \cite{bib53}
& CE + Dice
& CE + Dice
& CE + Dice
& Focal \cite{bib54} \\
\epochs{epochs} / \iters{iterations} & \epochs{10K} & \epochs{1000} & \epochs{1000} & \iters{30K} & \epochs{200} \\

\midrule
\textbf{Inference Strategy} &&&&& \\
ensemble
& 5-fold CV
& 5-fold CV
& n/a
& 2 folds of CV (space limit)
& n/a \\
post-processing
& breast bbox
& \begin{tabular}[c]{@{}c@{}}largest connected\\ component\end{tabular}
& n/a
& n/a
& breast bbox \\
\bottomrule
\end{tabular}
\begin{tablenotes}
\footnotesize
\item[a] For ensemble approaches, the number of parameters is the total parameter count over all the models in the ensemble.
Public implementations are available at the repositories provided by the participating teams: \url{https://github.com/jojoka-234/mama-mia-challenge}, \url{https://github.com/FraunhoferMEVIS/MAMA-MIA-Challange-FME}, \url{https://github.com/VICTORIA-project/MAMA-MIA_Challenge}, \url{https://github.com/Muhammad-Al-Barbary/MAMA-MIA_MartelLab}, \url{https://github.com/luvargas2/AIH-MAMA}
\end{tablenotes}
\end{threeparttable}
\label{tab2}
\end{table*}

\subsubsection{Task 1: Primary Tumor Segmentation}

The primary tumor segmentation task attracted a diverse set of approaches that, nevertheless, shared several common design principles. A comparative summary of the top five segmentation methods is provided in Table~\ref{tab2}. All approaches employ fully three-dimensional architectures operating directly on volumetric DCE-MRI data, reflecting the importance of spatial context for accurate tumor delineation. Input standardization through resampling and intensity normalization was consistently adopted, while voxel spacing, interpolation strategies, and normalization schemes varied across teams.

Most methods incorporated multiple DCE phases as input channels, typically including the pre-contrast and one or more early post-contrast acquisitions, in order to capture tumor enhancement dynamics. Architecturally, the majority of approaches were based on nnU-Net variants, including residual encoder configurations, while one method explored a 3D Vision Transformer backbone. Inference strategies frequently relied on ensembling across cross-validation folds, and several methods applied lightweight post-processing steps to suppress anatomically implausible segmentations.

\numone\textbf{ MIC}:
The MIC team proposed a residual-encoder nnU-Net with large-scale self-supervised pretraining on external DCE-MRI data~\cite{bib55}. A masked autoencoding strategy was used to pretrain on $n{=}4{,}799$ public DCE-MRI volumes, followed by fine-tuning on the MAMA-MIA dataset~\cite{bib31}. Multi-phase input and five-fold ensembling were combined with breast-bounding-box post-processing to improve robustness and subgroup consistency. Further details are reported in~\cite{bib56}.

\numtwo\textbf{ FME}
The FME team adopted a robustness-oriented strategy based on a single subtraction image (first post-contrast minus pre-contrast) to reduce sensitivity to acquisition variability. A five-fold ensemble of 3D residual-encoder nnU-Nets was trained, complemented by axis flipping test time augmentation and largest connected component filtering. This design prioritized generalization and stability across unseen centers.

\numthree\textbf{ ViCOROB}
%~\cite{EskreisWinkler2022BPE}
The ViCOROB team, as described in ~\cite{bib57}, employed a residual-encoder architecture using three representative DCE phases selected to preserve early enhancement kinetics. Breast-level cropping focused on the tumor-containing region, eliminating contralateral interference and improving signal-to-noise ratio. To enhance robustness across multi-center data, percentile-based intensity clipping, isotropic resampling, and z-score normalization were applied to standardize inputs.

\numfour\textbf{ Martel Lab}
The Martel Lab introduced a joint learning framework that simultaneously performs tumor segmentation and pCR prediction using a shared encoder. A 3D Vision Transformer\cite{bib58} pretrained on large-scale medical imaging data~\cite{bib59} was combined with temporal attention modules to model contrast enhancement dynamics across multiple DCE phases within an end-to-end multi-task architecture.

\numfive\textbf{ AIH-MAMA}
The AIH-MAMA team applied FairMedSeg~\cite{bib60}, a fairness-aware segmentation framework that integrates subgroup performance divergence directly into the training objective. Clinically relevant subgroups were identified using DivExplorer~\cite{bib61}, and sample reweighting was used to mitigate performance disparities without requiring subgroup metadata at inference time~\cite{bib62}.

\begin{table}[h]
\caption{Technical summary of top-performing pathologic complete response (pCR) prediction methods (Task 2). Abbreviations used in the table -  nnUNet pre: default nnUNet \cite{bib13} preprocessing strategy; bbox: Bounding Box; p0/p1/p2: pre-contrast, first post-contrast and second post-contrast phases; px: last available post-contrast phase; BCE: Binary Cross Entropy; CE: Cross Entropy; CV: Cross Validation; n/a: not applicable/available.}
\centering
\scriptsize

\setlength{\tabcolsep}{2pt}
\definecolor{rowgray}{gray}{0.93}
\rowcolors{3}{rowgray}{white}
\begin{threeparttable}
\begin{tabular}{lccc}
\toprule
 & \textbf{\numonec \text{ pimed-lab}} & \textbf{ \numtwoc  \text{ FME}} & \textbf{\numthreec  \text{ AI Strollers}} \\
\midrule

\textbf{Preprocessing} &&& \\
resample & $1 \times 1 \times 1$ & varying per model & \begin{tabular}[c]{@{}c@{}}nnUNet pre \cite{bib13} \\ $0.7 \times 0.7 \times 2.0$ \end{tabular}   \\
crop / pad & lesion bbox & lesion bbox & \xmark \\
reorient & \xmark & \xmark & \xmark \\
clip & \xmark & \xmark & \xmark\\
normalize &  z-score & z-score & z-score\\

\midrule
\textbf{Training Strategy} &&& \\

DCE phases & \begin{tabular}[c]{@{}c@{}} p1,..px (train) \\ p1 (inference) \end{tabular} & p0, p1, p2  & p0, p1, p2 \\

network\tnote{a} & \begin{tabular}[c]{@{}c@{}}3D ResNet-18 \cite{bib63}\\ mc3\_18\cite{bib64}\end{tabular} & \begin{tabular}[c]{@{}c@{}} r3d\_18, swin3d\_t, \\ r2plus1d\_18, mc3\_18 \cite{bib65} \end{tabular} & \begin{tabular}[c]{@{}c@{}}\epochs{SegResNet \cite{bib14}} \\ \iters{XGBoost \cite{bib66}} \end{tabular}\\

parameters (M)\tnote{b}
& 44.65
& 593.6
& 2.61 \\

pretrained & \xmark & \cmark & \xmark \\

optimizer & AdamW \cite{bib52} &  AdamW \cite{bib52} & 
\begin{tabular}[c]{@{}c@{}}\epochs{Adam \cite{bib67}} \\ \iters{n/a}\end{tabular} \\

batch size & 48 & [16, 32] & 
\begin{tabular}[c]{@{}c@{}}\epochs{2} \\ \iters{n/a}\end{tabular} \\

loss & CE + NT-Xent \cite{bib68} & BCE & 
\begin{tabular}[c]{@{}c@{}}\epochs{\begin{tabular}[c]{@{}c@{}} CE + Dice \\ + FRLoss \cite{bib69}\end{tabular}} \\ \iters{log-loss}\end{tabular} \\

epochs & 200 & 25 & 
\begin{tabular}[c]{@{}c@{}}\epochs{1000} \\ \iters{n/a}\end{tabular} \\
\midrule
\textbf{Inference Strategy} &&& \\
ensemble & 2 models & \begin{tabular}[c]{@{}c@{}}$5 \text{ models}\times 5\text{-fold CV}$\\(25 models)\end{tabular} & \begin{tabular}[c]{@{}c@{}}2 estimators\end{tabular}\\
\bottomrule
\end{tabular}
\begin{tablenotes}
\footnotesize
\item[a] AI Strollers combined two models: a segmentation model for feature extraction (black) and a pCR prediction model (gray).
\item[b] For ensemble approaches, the number of parameters is the total parameter count over all the models in the ensemble. Public implementations are available at the repositories provided by the participating teams: \url{https://github.com/dnschouten/mamamia_code_submission}, \url{https://github.com/FraunhoferMEVIS/MAMA-MIA-Challange-FME}, \url{https://github.com/slndaniel/MamaMiaSubmission}
\end{tablenotes}
\end{threeparttable}
\label{tab3}
\end{table}

\subsubsection{Task 2: Pathologic Complete Response Prediction}

For the pCR prediction task, participating methods explored complementary strategies for modeling treatment response from pre-treatment DCE-MRI. Table~\ref{tab3} summarizes the key technical characteristics of the top three classification approaches. All methods formulated pCR prediction as a supervised binary classification problem, but differed substantially in how imaging information was represented and exploited.

Two approaches adopted lesion-centered, end-to-end 3D classification pipelines, extracting fixed-size regions-of-interest around the primary tumor. These methods varied in their handling of temporal information, either treating multiple post-contrast phases as separate input channels or using phase selection and augmentation strategies. In contrast, one approach decoupled segmentation and classification by reusing deep features extracted from a pretrained segmentation network.

\numonec\textbf{ pimed-lab}
The pimed-lab team formulated pCR prediction as a lesion-centered 3D classification task and adopted a hybrid supervision strategy combining self-supervised learning through a 3D SimCLR~\cite{bib68} objective early in training, gradually transitioning to supervised learning once robust features had been learned. Two 3D convolutional architectures~\cite{bib63,bib64} were trained and ensembled to improve feature learning and generalization under limited labeled data.

\numtwoc\textbf{ FME}
The FME team employed pretrained 3D video classification architectures~\cite{bib65} operating on lesion-centered multi-phase DCE-MRI inputs. Hyperparameters were optimized using Ray Tune~\cite{bib70} via cross-validation, and model selection explicitly considered a composite score balancing predictive performance and subgroup fairness. A large ensemble and test time augmentation were used at inference to improve robustness.

\numthreec\textbf{ AI Strollers}
The AI Strollers team proposed a modular pipeline in which deep features were extracted from a pretrained segmentation network based on parameter-efficient SegRestNet~\cite{bib14}. These features were subsequently used to train an XGBoost classifier~\cite{bib66}, decoupling voxel-level segmentation learning from patient-level response prediction.

\begin{figure*}[h]
    \centering
    \begin{subfigure}{0.48\textwidth}
        \centering
        \includegraphics[width=\linewidth]{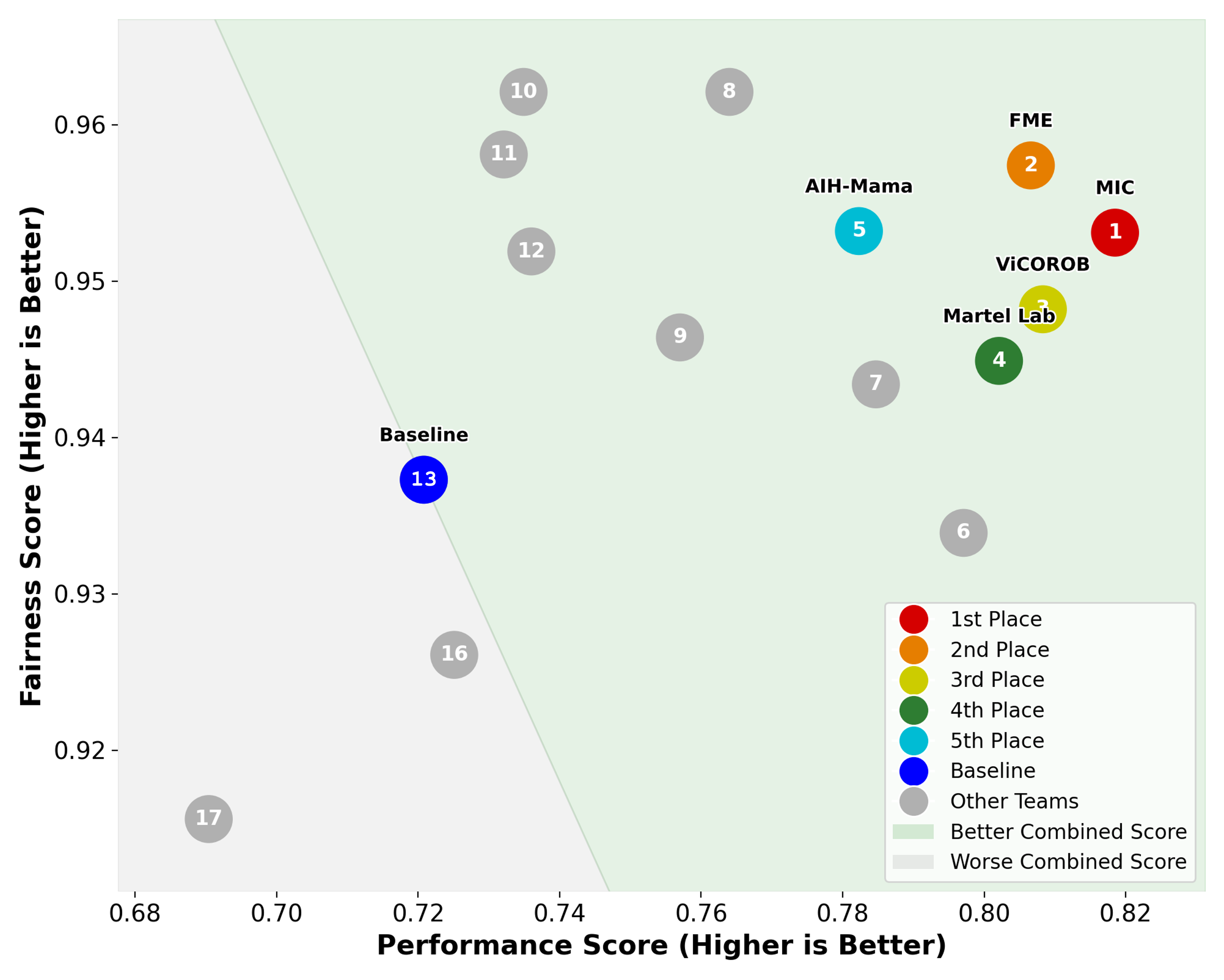}
        \caption{Primary Tumor Segmentation (Task 1).}
        \label{fig1a}
    \end{subfigure}
    \hfill
    \begin{subfigure}{0.48\textwidth}
        \centering
        \includegraphics[width=\linewidth]{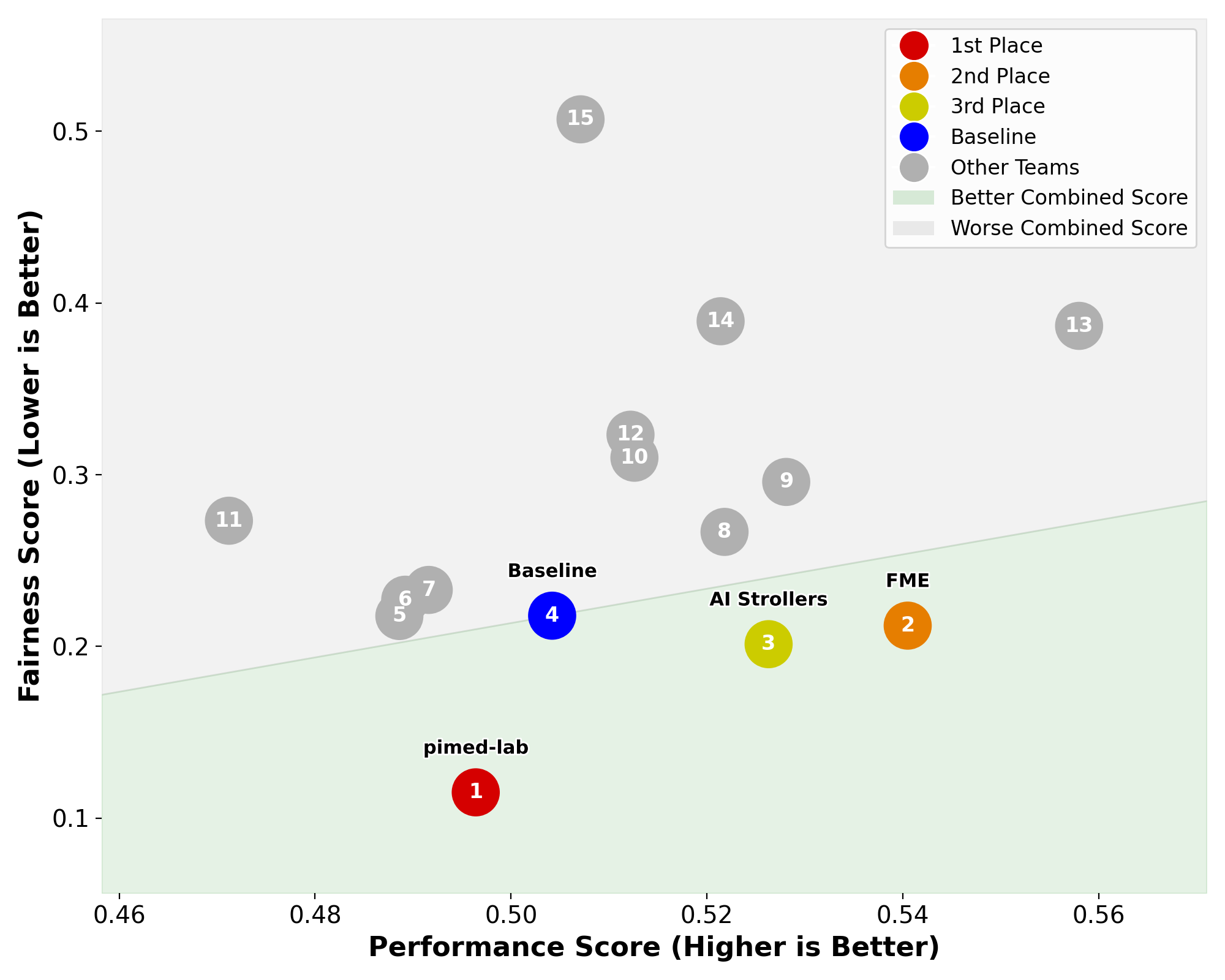}
        \caption{Pathologic Complete Response Prediction (Task 2).}
        \label{fig1b}
    \end{subfigure}
    \caption{Fairness plotted against performance across the two MAMA-MIA benchmark tasks. Teams with fairness-performance tradeoffs within the area indicated in green improve upon the baseline tradeoff. Note that the axis ranges are restricted to improve readability; consequently, teams falling outside these bounds are not shown.}
    \label{fig1}
\end{figure*}

\begin{figure*}
    \centering
    \includegraphics[width=0.9\linewidth]{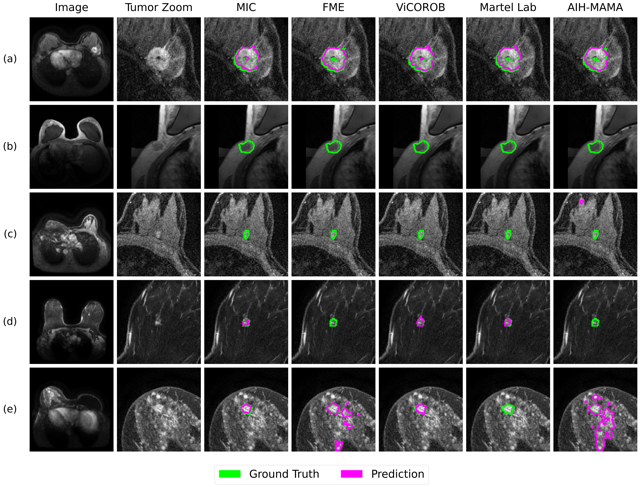}
    \caption{Qualitative results for challenging cases. The columns correspond to the MRI image, the zoomed-in tumor region (Tumor Zoom), and the top-5 team predictions (predicted segmentations) overlaid with the expert annotated tumor segmentation (ground truth). Results are shown for: (a) clip artifact (successful), (b) breast implant (failure), (c) low contrast (failure), (d) small tumor (high variance between teams), and (e) non-mass enhancement (high variance between teams). Artifacts, low discriminability, and atypical morphology remain key challenges.}
    \label{fig2}
\end{figure*}

\section{Results} \label{sec:results}
A total of 21 teams from 14 countries submitted valid entries to the final test phase of the benchmark. 13 teams submitted to both tasks, while 7 teams and 1 team submitted exclusively to Task 1 and Task 2, respectively. This section presents and analyzes the results obtained by the participating teams. As shown in Fig. \ref{fig1}, 12 teams outperformed the baseline in both fairness and performance for Task 1 (Fig. \ref{fig1a}). In Task 2 (Fig. \ref{fig1b}), only three teams demonstrated improvements over the baseline random model; all three improved fairness, while only two also achieved higher-than-baseline performance. The results for each task are presented in the following dedicated subsections.

\begin{table*}
\centering
\caption{Final leaderboard for Task~1: Primary Tumor Segmentation. Confidence intervals (95\%) for all Task 1 metrics were estimated using a non-parametric percentile bootstrap ($B=1000$, seed=42). DSC, NormHD, and Performance Score CIs were obtained by resampling patients with replacement and recomputing the mean metric on each bootstrap sample. Fairness Score CIs were computed using patient-level bootstrapping with repeated group-level disparity estimation across demographic groups. Combined Score CIs were derived jointly from the same bootstrap resamples for both performance and fairness components.}
\begin{threeparttable}
\scriptsize
\setlength{\tabcolsep}{3pt} 
\begin{tabular}{@{}cllllll@{}}
\toprule
\textbf{Rank} & \textbf{Team} &
\multicolumn{1}{c}{\textbf{\shortstack{Combined\\Score [95\% CI] $\uparrow$}}} &
\multicolumn{1}{c}{\textbf{\shortstack{Fairness\\Score [95\% CI] $\uparrow$}}} &
\multicolumn{1}{c}{\textbf{\shortstack{Performance\\Score [95\% CI] $\uparrow$}}} &
\multicolumn{1}{c}{\textbf{\shortstack{DSC\\{[95\% CI]} $\uparrow$}}} &
\multicolumn{1}{c}{\textbf{\shortstack{NormHD\\{[95\% CI]} $\downarrow$}}} \\
\midrule
1 & MIC &
0.8858 [0.8577, 0.8969] &
0.9531 [0.9051, 0.9651] &
0.8185 [0.8018, 0.8350] &
0.7360 [0.7160, 0.7539] &
0.0990 [0.0816, 0.1172] \\
2 & FME &
0.8820 [0.8464, 0.8913] &
0.9574 [0.8976, 0.9659] &
0.8066 [0.7873, 0.8248] &
0.7125 [0.6895, 0.7330] &
0.0993 [0.0815, 0.1182] \\
3 & ViCOROB &
0.8782 [0.8485, 0.8891] &
0.9482 [0.8963, 0.9643] &
0.8083 [0.7913, 0.8248] &
0.7182 [0.6976, 0.7381] &
0.1017 [0.0859, 0.1189] \\
4 & Martel Lab &
0.8735 [0.8471, 0.8856] &
0.9449 [0.8999, 0.9619] &
0.8021 [0.7863, 0.8182] &
0.7121 [0.6918, 0.7315] &
0.1078 [0.0915, 0.1244] \\
5 & AIH-Mama &
0.8677 [0.8379, 0.8781] &
0.9532 [0.8984, 0.9661] &
0.7823 [0.7634, 0.8012] &
0.6914 [0.6701, 0.7139]$^{*}$ &
0.1268 [0.1081, 0.1452] \\
6 & HWT@YCH &
0.8655 [0.8365, 0.8789] &
0.9339 [0.8843, 0.9514] &
0.7971 [0.7798, 0.8143] &
0.7080 [0.6872, 0.7273] &
0.1138 [0.0967, 0.1324] \\
7 & Flamingo &
0.8640 [0.8287, 0.8781] &
0.9434 [0.8798, 0.9593] &
0.7847 [0.7634, 0.8033] &
0.7033 [0.6807, 0.7245]$^{*}$ &
0.1338 [0.1120, 0.1576] \\
8 & CALADAN &
0.8631 [0.8239, 0.8683] &
0.9621 [0.8947, 0.9629] &
0.7640 [0.7412, 0.7862] &
0.7022 [0.6811, 0.7246]$^{*}$ &
0.1742 [0.1487, 0.2018] \\
9 & bigAI &
0.8517 [0.8189, 0.8626] &
0.9464 [0.8860, 0.9596] &
0.7570 [0.7360, 0.7766] &
0.6872 [0.6649, 0.7076]$^{*}$ &
0.1732 [0.1503, 0.1970] \\
10 & Shangqi,Gao@CAM &
0.8485 [0.8157, 0.8551] &
0.9621 [0.9025, 0.9655] &
0.7349 [0.7144, 0.7555] &
0.6101 [0.5836, 0.6359] &
0.1404 [0.1223, 0.1584] \\
11 & GK\_KI &
0.8451 [0.8060, 0.8512] &
0.9581 [0.8896, 0.9606] &
0.7321 [0.7074, 0.7552] &
0.6330 [0.6083, 0.6570] &
0.1688 [0.1441, 0.1955] \\
12 & Jeff &
0.8439 [0.8071, 0.8529] &
0.9519 [0.8886, 0.9570] &
0.7360 [0.7107, 0.7588] &
0.7025 [0.6819, 0.7221] &
0.2305 [0.1979, 0.2636]$^{*}$ \\
13 & \textit{Baseline} &
0.8290 [0.7868, 0.8416] &
0.9373 [0.8645, 0.9516] &
0.7208 [0.6984, 0.7432] &
0.6871 [0.6662, 0.7075] &
0.2455 [0.2125, 0.2806] \\
14 & Dynamo &
0.8290 [0.7868, 0.8416] &
0.9373 [0.8645, 0.9516] &
0.7208 [0.6984, 0.7432] &
0.6871 [0.6662, 0.7075]$^{*}$ &
0.2455 [0.2126, 0.2806]$^{*}$ \\
15 & PM &
0.8290 [0.7868, 0.8416] &
0.9373 [0.8645, 0.9516] &
0.7208 [0.6984, 0.7432] &
0.6871 [0.6662, 0.7075]$^{*}$ &
0.2455 [0.2125, 0.2806]$^{*}$ \\
16 & AEHRC-MIA &
0.8256 [0.7876, 0.8399] &
0.9261 [0.8605, 0.9405] &
0.7251 [0.7010, 0.7490] &
0.6781 [0.6562, 0.7004]$^{*}$ &
0.2280 [0.1960, 0.2601]$^{*}$ \\
17 & AI Strollers &
0.8030 [0.7507, 0.8249] &
0.9156 [0.8211, 0.9459] &
0.6904 [0.6612, 0.7189] &
0.6296 [0.6005, 0.6568] &
0.2489 [0.2166, 0.2818]$^{*}$ \\
18 & MedImgLab\_Unipa &
0.7270 [0.6912, 0.7438] &
0.9084 [0.8432, 0.9292] &
0.5456 [0.5208, 0.5716] &
0.4717 [0.4466, 0.4979] &
0.3805 [0.3494, 0.4110] \\
19 & FPixel &
0.7270 [0.6912, 0.7438] &
0.9084 [0.8432, 0.9292] &
0.5456 [0.5208, 0.5716] &
0.4717 [0.4466, 0.4979] &
0.3805 [0.3494, 0.4110] \\
20 & BWS-KNU &
0.7257 [0.6666, 0.7348] &
0.9382 [0.8352, 0.9368] &
0.5132 [0.4740, 0.5497] &
0.4556 [0.4245, 0.4859] &
0.4291 [0.3816, 0.4790] \\
21 & CIG@Illinois &
0.6593 [0.6122, 0.6802] &
0.8931 [0.8086, 0.9213] &
0.4256 [0.3968, 0.4550] &
0.5195 [0.4936, 0.5456] &
0.6683 [0.6265, 0.7096] \\
\bottomrule
\end{tabular}
\begin{tablenotes}
\footnotesize
\item[*] DSC and NormHD not statistically different from the baseline according to a paired $t$-test. All other values are significantly different ($p < 0.05$).
\end{tablenotes}
\end{threeparttable}
\label{tab4}
\end{table*}

\begin{table*}
\centering
\caption{Final leaderboard for Task~2: Treatment Response Prediction to Neoadjuvant Chemotherapy. Note that AUC is reported as an additional result and is excluded from the ranking scheme. Confidence intervals (95\% CI) were estimated using a non-parametric percentile bootstrap ($B=1000$, seed=42). Balanced Accuracy and AUC were bootstrapped from prediction–label pairs, while Fairness and Combined Scores used patient-level resampling with repeated equalized-odds disparity computation across demographic groups.  The Combined Score CI was obtained by repeating the resampling jointly for performance and fairness components.}
\begin{threeparttable}
\setlength{\tabcolsep}{10pt}
\scriptsize
\begin{tabular}{clcccc}
\toprule
\textbf{Rank} & \textbf{Team} &
\multicolumn{1}{c}{\textbf{\shortstack{Combined\\Score [95\% CI] $\uparrow$}}} &
\multicolumn{1}{c}{\textbf{\shortstack{Fairness\\Score [95\% CI] $\downarrow$}}} &
\multicolumn{1}{c}{\textbf{\shortstack{Performance\\Score [95\% CI] $\uparrow$}}} &
\multicolumn{1}{c}{\textbf{\shortstack{\cellcolor{gray!15}AUC [95\% CI]\\\cellcolor{gray!15}$\uparrow$}}} \\
\midrule
1 & pimed-lab &
0.6907 [0.6364, 0.7272] &
0.1150 [0.0399, 0.2292] &
0.4964 [0.4853, 0.5102] &
\cellcolor{gray!15}0.4556 [0.4035, 0.5121] \\
2 & FME &
0.6642 [0.4893, 0.6602] &
0.2121 [0.2232, 0.5561] &
0.5405 [0.4992, 0.5841]$^{+}$ &
\cellcolor{gray!15}0.5722 [0.5195, 0.6305] \\
3 & AI Strollers &
0.6625 [0.4778, 0.6569] &
0.2013 [0.2148, 0.5663] &
0.5263 [0.4870, 0.5658] &
\cellcolor{gray!15}0.5218 [0.4629, 0.5771] \\
\multicolumn{1}{c}{4} & \textit{Baseline} &
0.6431 [0.4600, 0.6360] &
0.2179 [0.2418, 0.5701] &
0.5042 [0.4551, 0.5498] &
\cellcolor{gray!15}0.5031 [0.4455, 0.5551] \\
5 & BWS-KNU &
0.6355 [0.4792, 0.6325] &
0.2177 [0.2205, 0.5324] &
0.4886 [0.4478, 0.5314] &
\cellcolor{gray!15}0.5075 [0.4513, 0.5656] \\
6 & Martel Lab &
0.6310 [0.4802, 0.6316] &
0.2272 [0.2397, 0.5164] &
0.4892 [0.4477, 0.5316] &
\cellcolor{gray!15}0.4740 [0.4195, 0.5330] \\
7 & Flamingo &
0.6293 [0.4459, 0.6238] &
0.2329 [0.2584, 0.5906] &
0.4916 [0.4449, 0.5384] &
\cellcolor{gray!15}0.4683 [0.4151, 0.5284] \\
8 & AEHRC-MIA &
0.6275 [0.4681, 0.6206] &
0.2667 [0.2889, 0.5832] &
0.5218 [0.4783, 0.5698] &
\cellcolor{gray!15}0.5378 [0.4819, 0.5996] \\
9 & Jeff &
0.6162 [0.4345, 0.6273] &
0.2958 [0.2815, 0.6459] &
0.5281 [0.4793, 0.5748] &
\cellcolor{gray!15}0.5075 [0.4519, 0.5633] \\
10 & MIC &
0.6013 [0.5046, 0.6348] &
0.3099 [0.2448, 0.5035] &
0.5126 [0.4716, 0.5530] &
\cellcolor{gray!15}0.5000 [0.5000, 0.5000] \\
11 & CALADAN &
0.5990 [0.4363, 0.5978] &
0.2732 [0.2818, 0.5951] &
0.4712 [0.4243, 0.5173] &
\cellcolor{gray!15}0.4819 [0.4264, 0.5417] \\
12 & HWT@YCH &
0.5945 [0.4282, 0.6186] &
0.3233 [0.2790, 0.6543] &
0.5122 [0.4643, 0.5591] &
\cellcolor{gray!15}0.5193 [0.4650, 0.5704] \\
13 & PM &
0.5856 [0.4407, 0.6117] &
0.3867 [0.3432, 0.6739] &
0.5580 [0.5111, 0.6014]$^{*}$ &
\cellcolor{gray!15}0.5702 [0.5160, 0.6221] \\
14 & ViCOROB &
0.5660 [0.4608, 0.6101] &
0.3894 [0.2989, 0.6001] &
0.5214 [0.4783, 0.5647] &
\cellcolor{gray!15}0.5595 [0.5036, 0.6151] \\
15 & bigAI &
0.5001 [0.3610, 0.5398] &
0.5070 [0.4317, 0.7725] &
0.5071 [0.4609, 0.5522] &
\cellcolor{gray!15}0.5011 [0.4461, 0.5528] \\
\bottomrule
\end{tabular}
\begin{tablenotes}
\footnotesize
\item[*] Balanced accuracy significantly higher than a random classifier based on 1000 random simulations ($p < 0.05$).
\item[+] Marginally higher than random ($0.05 \leq p < 0.10$).
\item[] All other methods showed no statistically significant difference from random ($p \geq 0.10$).
\end{tablenotes}
\end{threeparttable}
\label{tab5}
\end{table*}
\subsection{Task 1: Primary Tumor Segmentation}

Table \ref{tab4} summarizes final rankings. The top five teams improved over the baseline by 0.43–4.89\% DSC, with concurrent fairness gains ($\sim2\%$), indicating consistent performance improvements across subgroups.

Qualitative analysis (Fig. \ref{fig2}) shows strong performance for well-circumscribed mass lesions, while failures are concentrated in small, non-mass, low-contrast tumors and implant-associated artifacts. These cases exhibit high inter-method variability, indicating sensitivity to ambiguous morphology.

\begin{figure*}
    \centering
    \includegraphics[width=\linewidth]{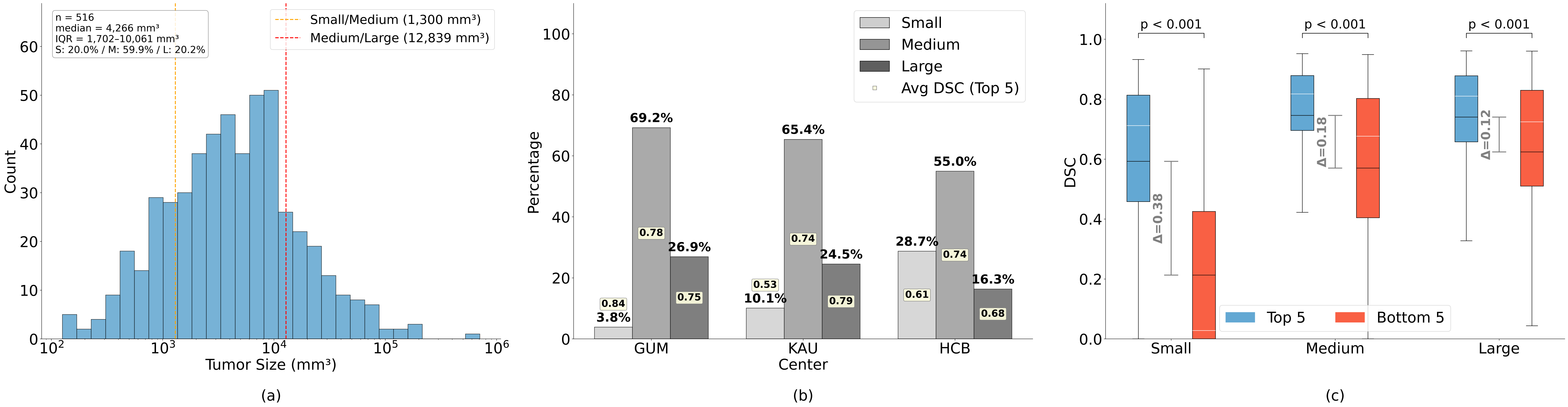}
    \caption{Impact of tumor size on segmentation performance. (a) Histogram of tumor volumes
($mm^{3}$) on a logarithmic scale, with dashed lines indicating the 20th and 80th percentile thresholds used to categorize tumors
as Small, Medium, or Large, (b) Percentage distribution of tumor size categories across the three participating centers (GUM, KAU, HCB), with boxed values showing the average Dice Similarity Coefficient (DSC) for the top 5 teams in each category. (c) Comparison of DSC distributions between the top 5 and bottom 5 performing teams across tumor size categories, with gap bars ($\Delta$) indicating the difference in mean DSC.  Brackets give two-sided Wilcoxon signed-rank p-values comparing the Top 5 against the Bottom 5 teams. For each patient, the mean DSC across the five top-ranked teams was compared with the mean DSC across the five bottom-ranked teams, so patients are the unit of analysis and the two arms are paired on identical cases.}
    \label{fig3}
\end{figure*}

\begin{figure}
    \centering    \includegraphics[width=\columnwidth]{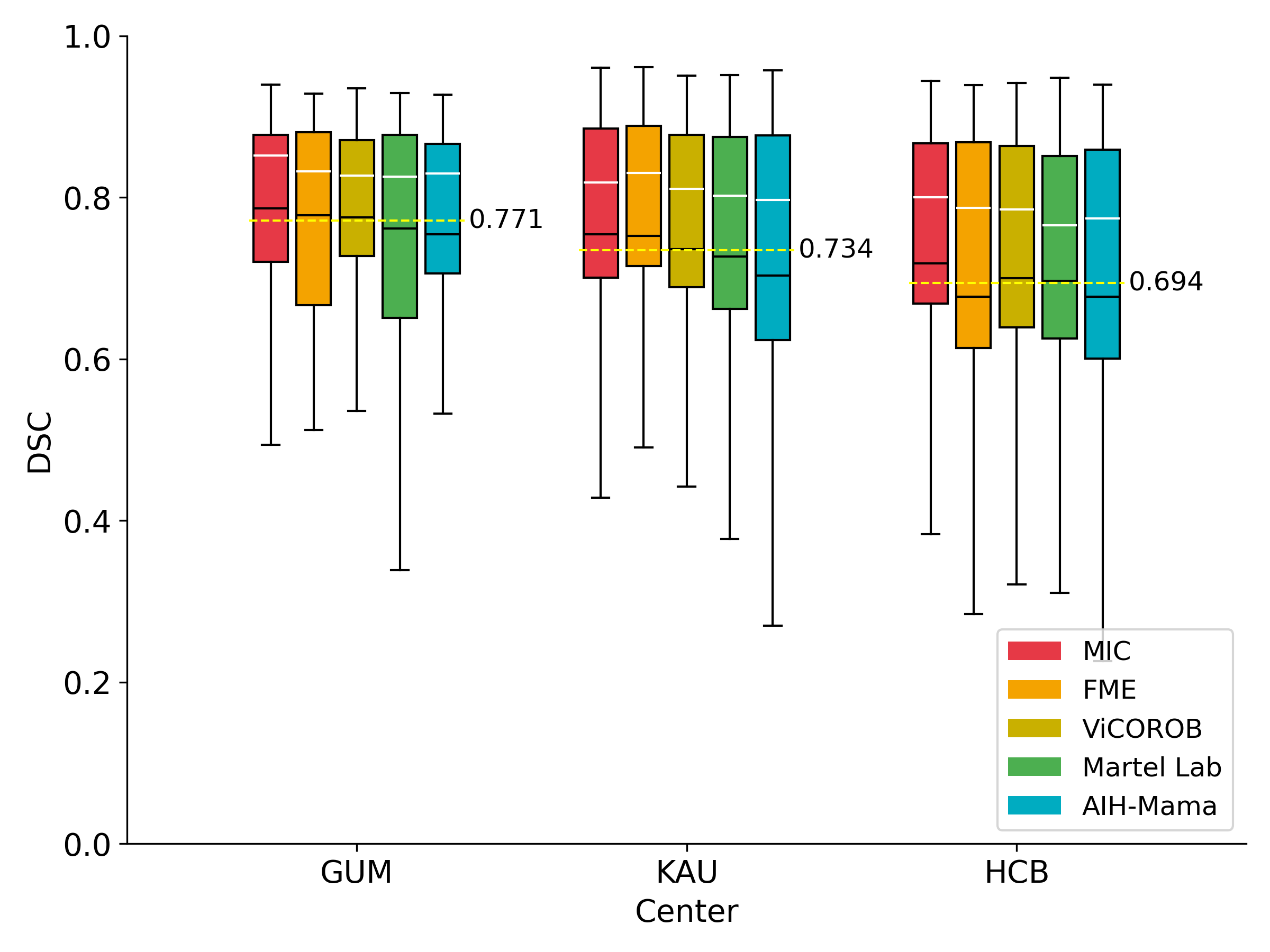}
    \caption{
Distribution of DSC across the three participating centers (GUM, KAU, HCB) for the top 5 performing teams in Task 1. Each boxplot represents the per-case DSC distribution for a team, in order of their ranking. Horizontal dashed lines indicate the average DSC across all top 5 teams for each center.}
    \label{fig4}
\end{figure}

This variability is consistent with the analysis shown in Fig. \ref{fig3}, which indicates that smaller tumors achieve lower average DSC compared to medium and large tumors. Notably, the performance gap between the top five and bottom five methods is largest for small tumors, suggesting that top-performing approaches yield the greatest relative improvements in this particularly challenging regime. Tumor size also influences performance across centers. Overall DSC varies by center, with the highest performance observed for GUM, followed by KAU and HCB, as shown in Fig. \ref{fig4}. This trend may be partly explained by differences in tumor size distributions across centers: centers with a higher proportion of small tumors, most notably KAU and HCB, tend to exhibit lower DSC. 

\begin{figure*}
\centering
\begin{subfigure}{0.4\linewidth}
    \centering
    \includegraphics[width=\linewidth]{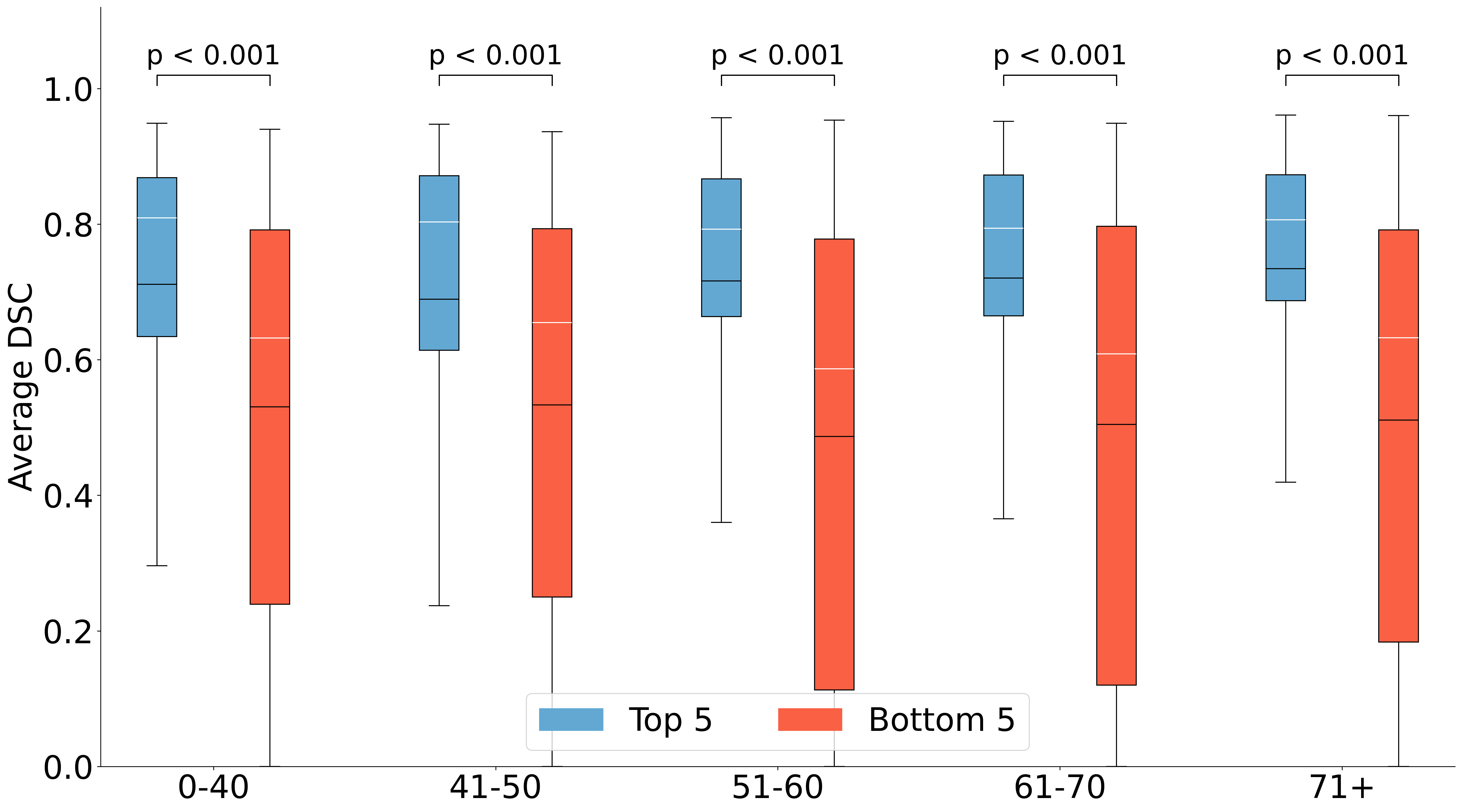}
    \caption{Age}
\end{subfigure}
\hspace{5em}
\begin{subfigure}{0.4\linewidth}
    \centering
    \includegraphics[width=\linewidth]{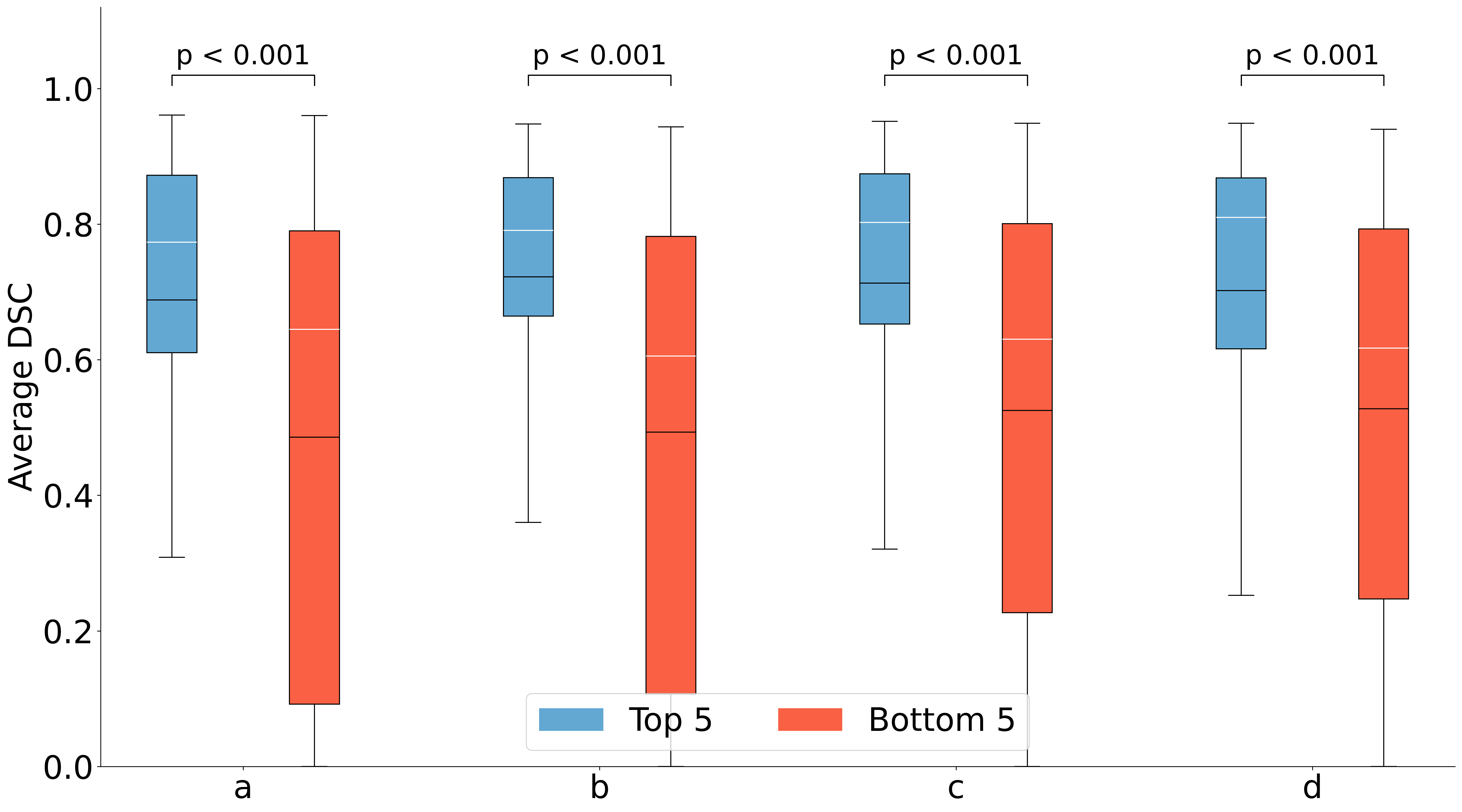}
    \caption{Breast Density}
\end{subfigure}
\begin{subfigure}{0.4\linewidth}
    \centering
    \includegraphics[width=\linewidth]{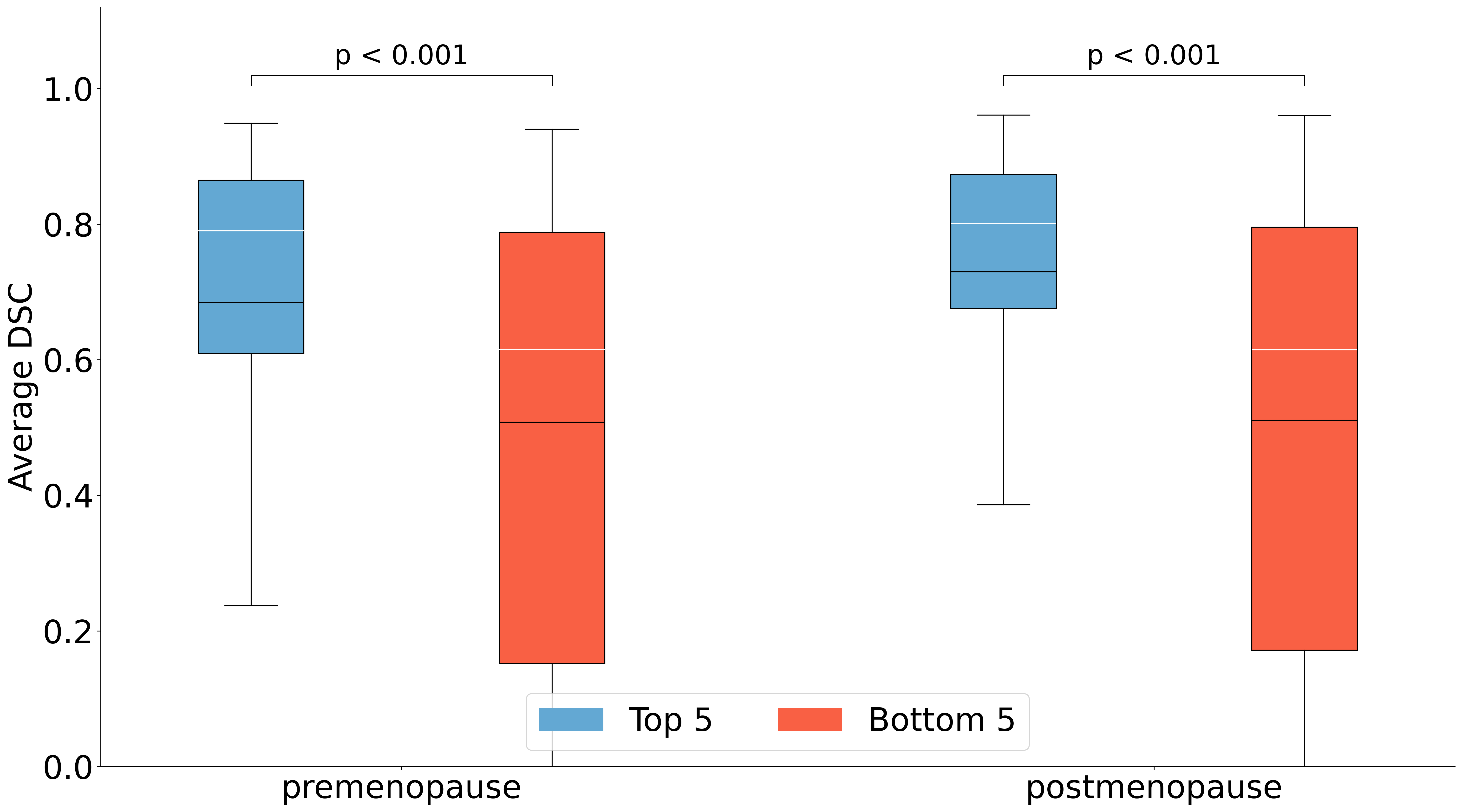}
    \caption{Menopausal Status}
\end{subfigure}
\hspace{5em}
\begin{subfigure}{0.4\linewidth}
    \centering
    \includegraphics[width=\linewidth]{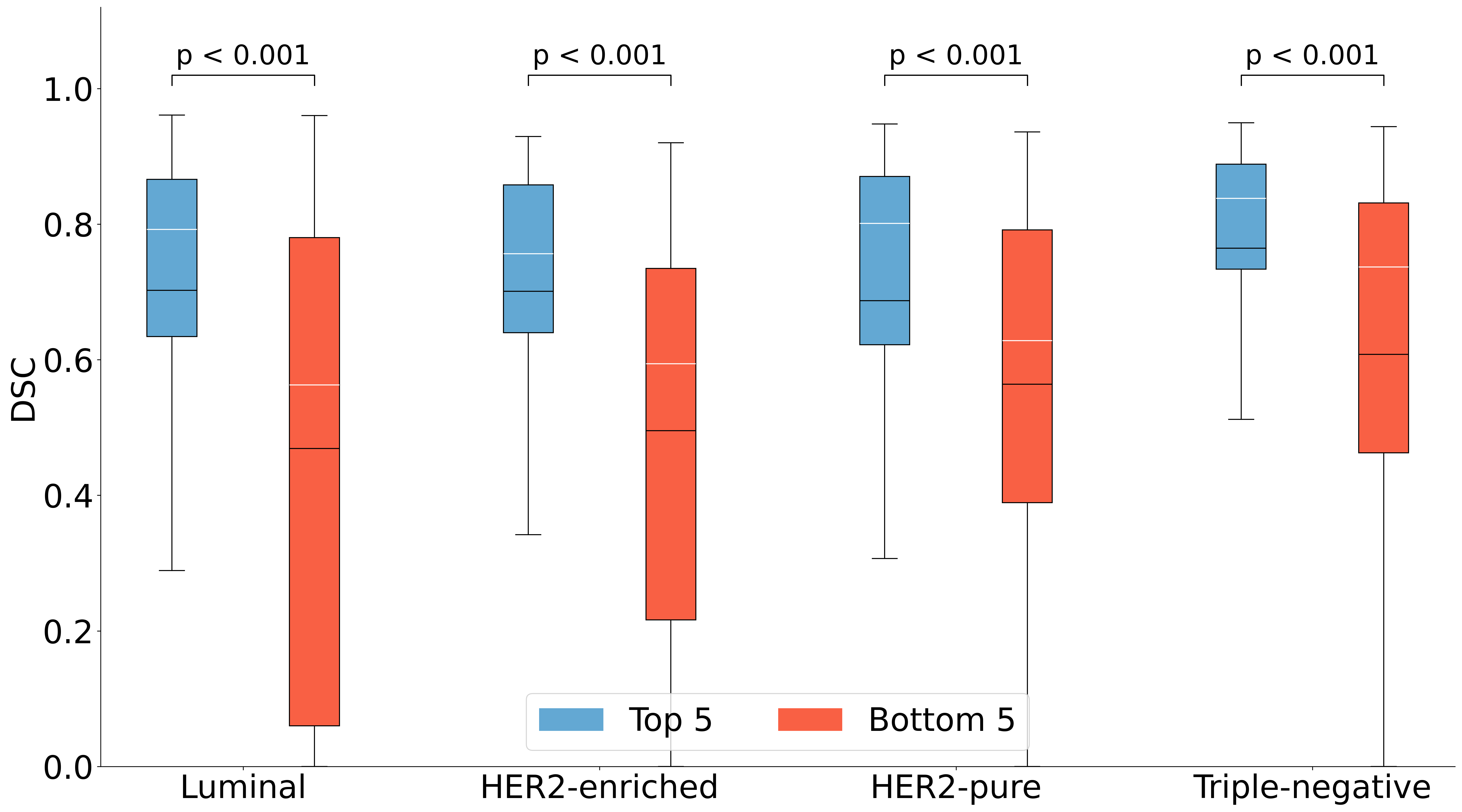}
    \caption{Tumor Subtype}
    \label{fig5d}
\end{subfigure}
\caption{Subgroup analysis of segmentation performance. Boxplots show DSC distributions for the top 5 (blue) and bottom 5 (red) performing teams, stratified by (a) breast density, (b) age group, (c) menopausal status, and (d) Tumor Subtype. Brackets give two-sided Wilcoxon signed-rank p-values comparing the Top 5 against the Bottom 5 teams. For each patient, the mean DSC across the five top-ranked teams was compared with the mean DSC across the five bottom-ranked teams, so patients are the unit of analysis and the two arms are paired on identical cases. The performance of high-ranking methods consistently improve in comparison to the low-ranking methods. There is no substantial differences between subgroups in the challenge fairness categories. Across tumor subtype categories, the triple\_negative subtype has the highest overall median DSC, while luminal category is especially challenging for the weaker models.}
\label{fig5}
\end{figure*}

The fairness analysis shown in Fig. \ref{fig5} demonstrates that the top five teams consistently achieved higher average DSC than the bottom five across all subgroups. Performance was largely comparable across subgroups defined by breast density, age, and menopausal status, with no systematic subgroup-specific differences observed. This suggests a high degree of fairness and robustness in the analyzed methods with respect to these patient characteristics. In addition to the fairness variables, Fig. \ref{fig5d} compares segmentation performance across four different breast cancer tumor subtypes. The triple-negative subtype yields the best results, with the top 5 models achieving the highest mean DSC (0.76) and the tightest performance distribution. This suggests that tumors of this specific subtype might have the most distinct or recognizable morphological characteristics in the imaging modality used. Conversely, the luminal subtype proved to be the most challenging overall; it resulted in the lowest mean DSC (0.47) and the widest spread for the weaker models, while also producing the lowest individual outlier (DSC < 0.30) even among the top-performing methods.

We evaluated the stability of the segmentation leaderboard across varying levels of the fairness weighting parameter $\lambda$ (Fig. \ref{fig6a}). Because the segmentation models achieved relatively high overall performance, the final rankings proved resilient to adjustments in $\lambda$. When evaluating only based on performance score ($\lambda=1$), the the top four teams remain unchanged. Notably, the team AIH-MAMA, which incorporates explicit fairness adjustments into their algorithm, demonstrated significantly stronger fairness metrics than the teams in 5th and 6th position on performance score only ranking. Consequently, this strong fairness profile secures them the 5th position in the final aggregated leaderboard.

\subsection{Task 2: Pathologic Complete Response Prediction}

Table~\ref{tab5} summarizes the final rankings for pCR prediction. The top three teams achieved combined scores of 0.6907, 0.6642, and 0.6625, compared to 0.6431 for the baseline random model. Notably, top three submissions did not substantially outperform the baseline in performance score alone.

Only one team (PM) achieved performance significantly different from random prediction ($p=0.0140$), while FME showed marginal significance ($p=0.063$). All other methods were statistically indistinguishable from random performance ($p>0.1$), underscoring the intrinsic difficulty of pCR prediction from pre-treatment MRI alone.

Clear performance--fairness trade-offs were observed. While PM obtained the highest performance score (0.5580), it exhibited the largest fairness gap (0.3867), resulting in a lower combined score (rank 13). In contrast, the top-ranked teams achieved competitive performance with improved fairness, indicating balanced subgroup behavior. However, further analysis between subgroups is not conducted for this task due to the lack of discriminative performance by the submitted models.

In all submissions, recall for the pCR class was generally low (Fig. \ref{fig7}). For instance, the top-performing method, pimed-lab, achieved a pCR recall of just 0.01 compared to a 'not pCR' recall of 0.98 ($\Delta$ = +0.96). Similarly, FME and AI Strollers demonstrated low pCR recalls of 0.26, alongside much higher 'not pCR' recalls of 0.82 and 0.80, respectively ($\Delta$ = +0.56 and $\Delta$= +0.54). This suggests conservative prediction strategies and limited sensitivity to complete responders. This likely reflects both class imbalance and the inherent heterogeneity of treatment response patterns.

Analysis of the calibration curves for the top three methods (Fig. \ref{fig8}) reveals limited  separation in the predicted probabilities, which are predominantly clustered around the overall class prevalence, of approximately 0.27. The mean predicted probabilities are narrowly constrained; pimed-lab's predictions range only from $0.12$ to $0.45$, while FME and AI Strollers show slightly wider dispersions from $0.21$ to $0.59$ and $0.07$ to $0.70$, respectively. Furthermore, while the models output varying predicted probabilities, the actual fraction of true positives (pCR cases) fails to scale proportionally. For instance, even when AI Strollers and FME output their most confident predictions (0.70 and 0.59, respectively), the actual fraction of pCR cases in those subsets reaches only 0.33 and 0.38. For pimed-lab, the true positive rate never exceeds 0.35 despite predictions reaching 0.45. Because the actual response rates rarely rise meaningfully above the population average, all curves fall heavily below the reference line at higher thresholds, revealing systematic overconfidence and a shared inability to reliably identify true treatment responders.

\begin{figure*}[!t]
    \centering
    \begin{subfigure}{0.5\textwidth}
        \centering
        \includegraphics[width=\linewidth]{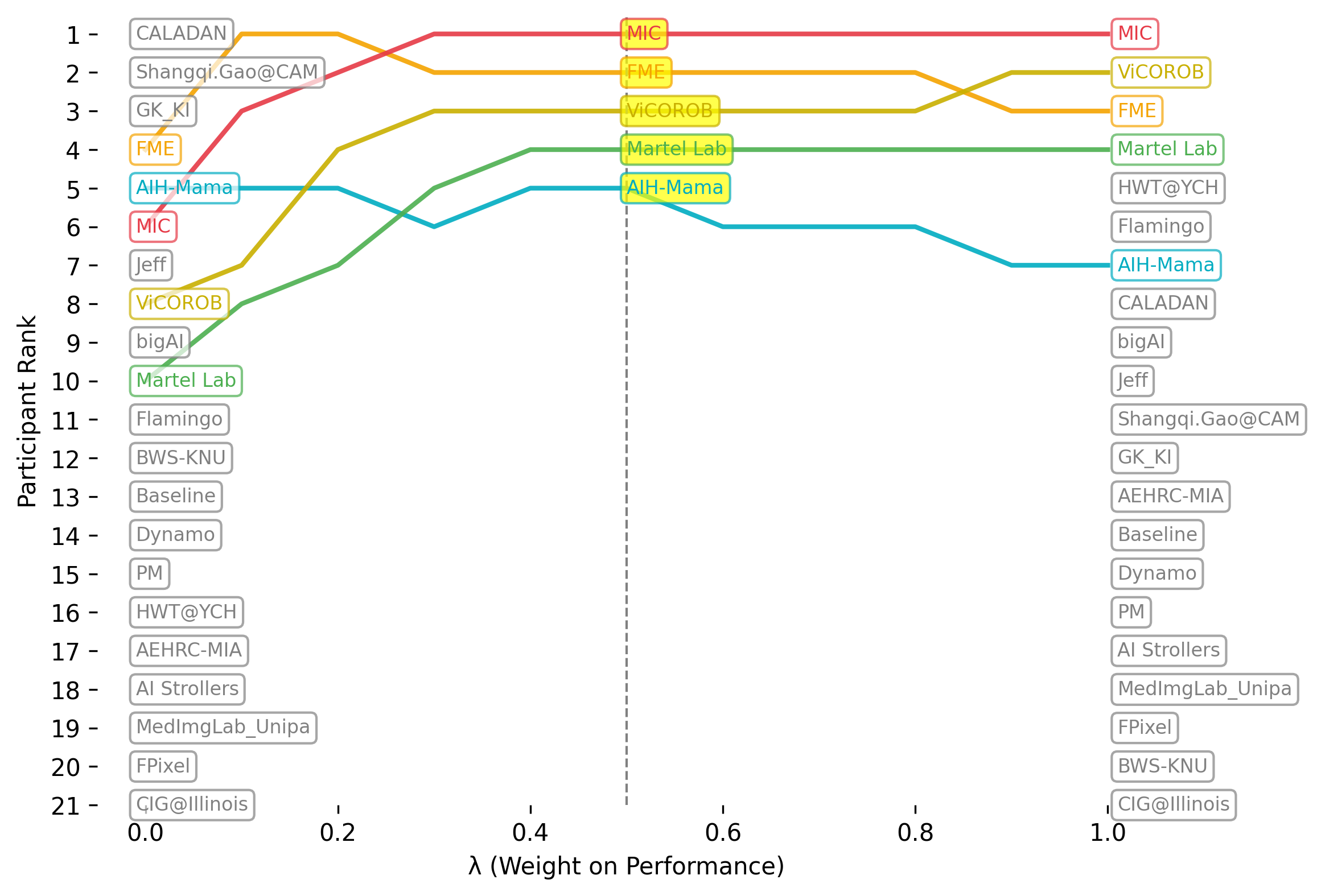}
        \caption{Task 1: $\lambda$ in Primary Tumor Segmentation}
        \label{fig6a}
    \end{subfigure}
    \hfill
    \begin{subfigure}{0.45\textwidth}
        \centering
        \includegraphics[width=\linewidth]{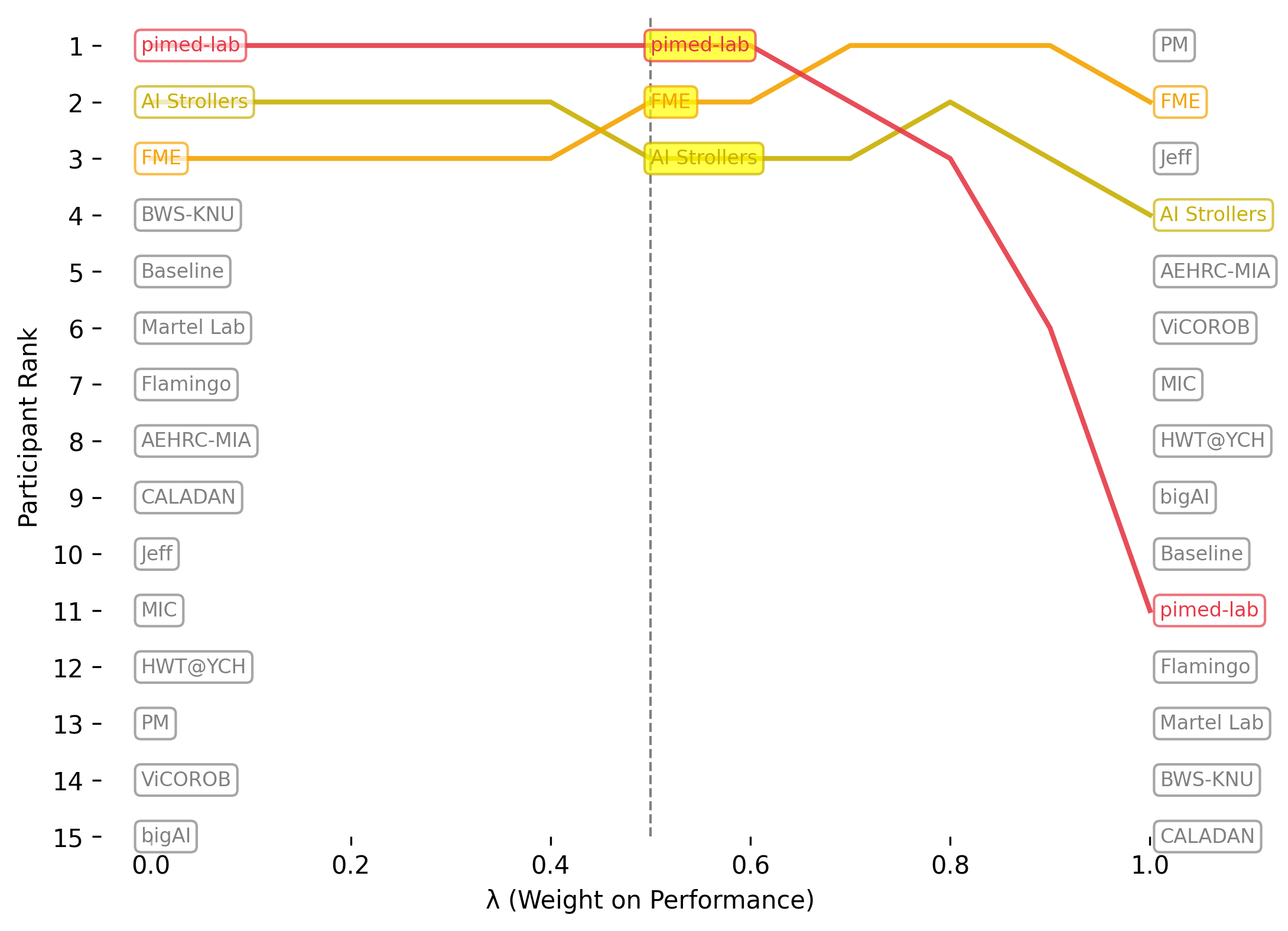}
        \caption{Task 2: $\lambda$ in Pathologic Complete Response Prediction}
        \label{fig6b}
    \end{subfigure}
    \caption{Fairness-versus-performance evaluation tradeoffs: Effect of $\lambda$ on the ranking across tasks and participating teams.}
    \label{fig6}
\end{figure*}

\begin{figure}
    \centering
    \includegraphics[width=\linewidth]{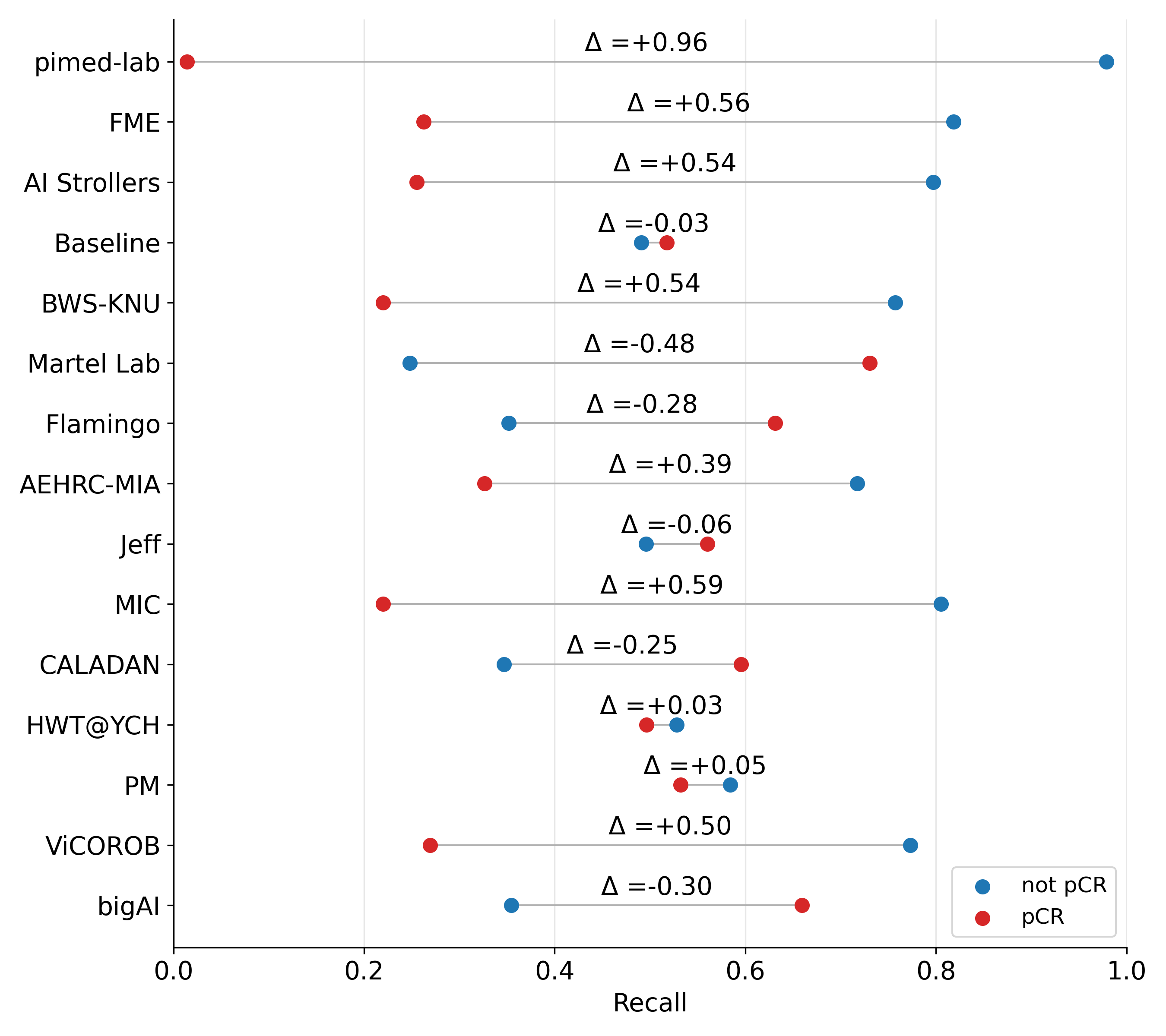}
    \caption{Recall comparison between pCR (minority) and not pCR (majority) classes across participating teams. $\Delta$ denotes the recall gap, reflecting class-wise prediction bias.}
    \label{fig7}
\end{figure}

\begin{figure*}[h]
    \centering
    \includegraphics[width=\linewidth]{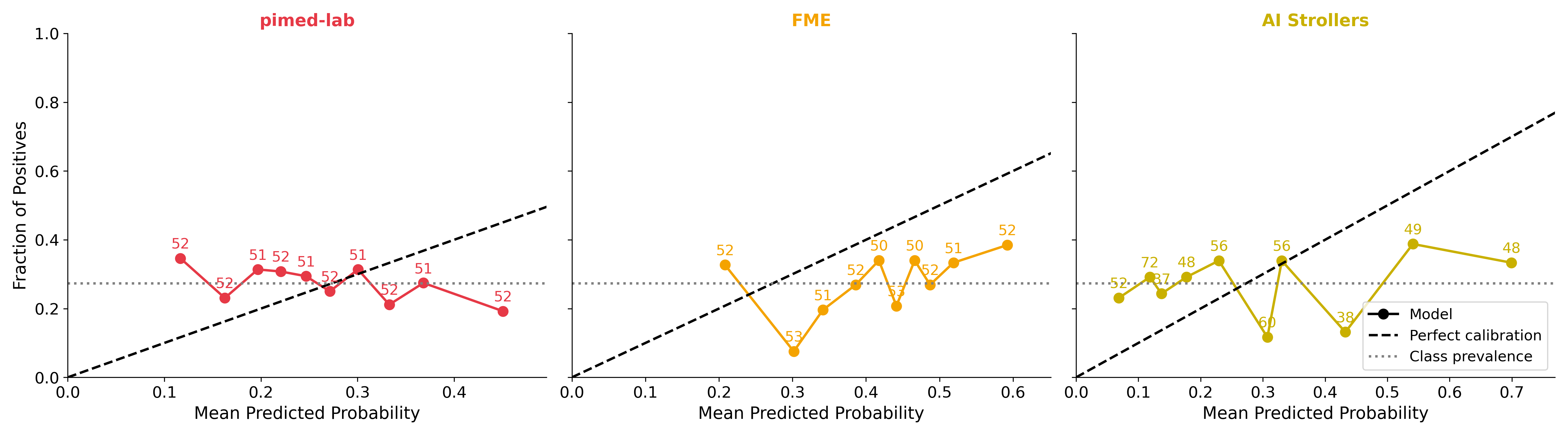}
    \caption{Reliability diagrams for top 3 teams in pCR prediction. Predicted probabilities are grouped into quantile bins; the fraction of true positives per bin is plotted against mean predicted probability. Dashed line: perfect calibration; dotted line: class prevalence. Annotated values indicate bin sample sizes.}
    \label{fig8}
\end{figure*}

We evaluated the stability of the classification leaderboard by analyzing how final rankings changed across varying levels of the fairness weighting parameter $\lambda$ (Fig. \ref{fig6b}). Because the pCR prediction task yielded lower overall performance scores, the rankings were highly sensitive to changes in $\lambda$. Setting $\lambda=1.0$ (i.e., ignoring fairness) results in participant PM moving from rank 13 out of 15 to rank~1, while only one (FME) of the top three teams at $\lambda=0.5$ remains in the top tier. Even modest increases in $\lambda$ lead to substantial reordering: at $\lambda=0.6$, 6 out of 15 teams change rank, and at $\lambda=0.8$, 13 out of 15 teams are affected.

\section{Discussion and Future Work} \label{sec:Discussion}

Building upon recent advances in breast MRI AI, including increasingly sophisticated externally validated segmentation and treatment-response prediction models, the MAMA-MIA Challenge was designed as a real-world stress test for AI in breast MRI, explicitly evaluating generalization to unseen clinical sites and consistency across patient subgroups under a common benchmark framework. We synthesize outcomes across primary tumor segmentation and pretreatment pCR prediction, analyzing performance trends, failure modes, and fairness behavior. As MAMA-MIA evolves from a one-time competition into a standardized benchmark with unified accuracy–fairness evaluation, we discuss implications for evaluation, clinical translation, and future research aimed at improving robustness and clinical relevance.

\subsection{Robustness of Primary Tumor Segmentation}

Primary tumor segmentation attracted 20 valid submissions to the final test phase, reflecting both its clinical relevance and methodological maturity. Under cross-continental evaluation, most top-performing methods improved over the nnU-Net baseline in both accuracy and subgroup consistency, indicating that contemporary 3D pipelines generalize reasonably well to unseen centers and heterogeneous acquisition protocols.

Methodologically, submissions showed strong convergence toward robustness-oriented design choices. All top five methods relied on fully 3D architectures, predominantly nnU-Net variants, with one exception adopting a 3D Vision Transformer backbone (Table~II). Ensemble-based inference was employed by four of the top five teams, together with standardized resampling and intensity normalization, highlighting the importance of variance reduction under cross-center deployment. With respect to input representation, four of the five top-ranked methods incorporated multiple DCE phases to explicitly model contrast enhancement dynamics, whereas only one method (e.g., FME) relied exclusively on a single subtraction image. While both strategies yielded competitive performance, no input design consistently mitigated failures in small, non-mass enhancement (NME), or low-contrast lesions, suggesting that current gains stem from incremental robustness improvements rather than fundamentally distinct paradigms.

Performance varied with tumor size (Fig. \ref{fig3}), with small tumors remaining challenging even for top-performing methods and accounting for the largest gap between top- and bottom-ranked teams (mean DSC delta 0.38 between averaged top 5 and bottom 5). Stability increased with tumor size both within and across performance tiers (e.g., mean DSC delta 0.12 for large tumors), indicating that leaderboard separation is largely driven by a method's ability to handle small, fragmented, or visually ambiguous lesions rather than larger cases.

Qualitative analysis (Fig. \ref{fig2}) corroborates these failure modes, including NME, weak lesion–background contrast, implant-related artifacts, and small or fragmented lesions. Substantial inter-method variability in such cases highlights model design trade-offs and the limitations of voxel-level overlap metrics in capturing clinically relevant uncertainty. Together, these findings indicate readiness for multi-center deployment in common scenarios, while motivating uncertainty-aware or interactive strategies for challenging lesions. 

Stratifying segmentation performance by molecular subtype revealed an interesting pattern (Fig. \ref{fig5d}): triple-negative tumors were segmented most accurately, whereas the luminal subtype was the most challenging, with the performance differences being more pronounced in bottom-5 methods. This ordering can be explained by subtype-specific imaging morphology observed in the literature. Specifically, triple-negative cancers predominantly present as round or oval, circumscribed, rim-enhancing masses~\cite{bib71}, producing high-contrast, and closed boundaries that the networks likely have an easier time delineating reliably. Luminal subtype is known to present as irregular tumors with spiculated margins, heterogenous enhancements, and in some cases, multifocal presentation~\cite{bib72}, which can make it a more difficult segmentation target. Further comparing these results to existing state-of-the-methods in breast tumor segmentation, a subtype-robustness analyses conducted on the MAMA-MIA dataset~\cite{bib73} reported overall lower dice coefficient values (0.741 -- 0.779), with the lowest performance also obtained on the luminal subtype. Another recent multicenter framework by Wang et al.~\cite{bib7} reports dice scores (n$=$687; subtype-wise DSC 0.820 -- 0.857) with similar values as the top-5 methods, however, the worst performance is reported on HER2-enriched subtype. The absence of tumor-wise labels (e.g. mass, non-mass, heterogenous, irregular etc.) prevents us from further analyzing these correlations, but it does make an interesting future direction.

\subsection{Challenges in Pretreatment pCR Prediction}

Pretreatment pCR prediction task from baseline DCE-MRI proved substantially more challenging than tumor segmentation in this challenge. Across 14 submissions, performance gains over the random baseline were marginal, rarely statistically significant, and often driven by fairness rather than discriminative accuracy (Table \ref{tab5}). Most models struggled with class imbalance and asymmetric recall, frequently defaulting to majority-class predictions while failing to reliably identify true pCR cases. Calibration analysis (Fig. \ref{fig8}) revealed limited probability separation and systematic overconfidence, suggesting reliance on weak imaging cues rather than robust response-related biomarkers.

FME was a consistently strong-performing submission across all evaluation settings, achieving the highest AUC among participating methods. Their approach combined an ensemble of pretrained CNN- and transformer-based models and test-time augmentation. Although the individual contributions of these components cannot be disentangled, the observed results suggest that such strategies may promote more robust representations, contributing not only to their predictive performance but also to the stability of the method across different values of the fairness weighting parameter $\lambda$. This approach contrasts with the findings from Martel Lab, who employed a vision transformer pretrained on a large-scale medical database for joint segmentation and pCR prediction. While it reached the top-5 in tumor segmentation, it did not consistently improve pCR prediction (AUC 0.47), indicating that representation scale alone may beinsufficient without complementary strategies for variance reduction.

The failure of highly accurate segmentation models to effectively predict pCR was not unique to Martel Lab. Participating approaches included lesion-centered end-to-end classifiers and pipelines leveraging segmentation-derived features (MIC team top-1 Task 1~\cite{bib56}), yet no clear relationship emerged between segmentation quality and pCR performance. While one team (AI Strollers) reused Task~1 features within an XGBoost classifier, the top three submissions on Task~2 generally addressed segmentation and response prediction as separate modeling problems, without employing a unified multi-task framework with shared feature representations. 

In this setup of predicting pCR from pre-treatment imaging, temporal information was restricted to contrast dynamics within DCE acquisition. All top-ranked methods incorporated multiple contrast phases (Table~\ref{tab3}), but differed in modeling strategy: two applied end-to-end deep learning to multi-phase inputs, while one used segmentation-derived features followed by a separate classifier, resulting in only implicit temporal modeling.

In regards to the fairness analysis, no clear and reliable trends emerged in Task 2 due to the overall difficulty of predicting pCR. Molecular subtype, which was used as an independent variable to analyze Task 1, is an intrinsic disease characteristic that strongly influences the likelihood of achieving pCR~\cite{bib74}. Consequently, treating subtype as a protected attribute would conflate a genuine biological determinant of the target outcome with an equity concern. Large clinical studies have consistently demonstrated substantial differences in pCR rates across subtypes, with luminal tumors exhibiting the lowest response rates and HER2-positive and triple-negative tumors achieving markedly higher rates~\cite{bib44}. Similarly, previous MRI studies have reported subtype-dependent differences in the diagnostic performance of imaging for treatment response assessment~\cite{bib75}, as well as improvement when subtype information is incorporated in the model \cite{bib76}.

Overall, the results show that a substantial gap remains between the promising metrics reported in the pCR prediction literature and the performance observed under a common benchmark setting. Several recent studies have demonstrated encouraging performance using pretreatment MRI with advanced imaging biomarkers and external validation cohorts \cite{bib8,bib9,bib10,bib11}. However, other multicenter studies have reached more cautious conclusions, reporting that pretreatment MRI radiomics alone provides limited predictive value, does not consistently outperform clinicopathologic models, or offers little additional benefit when combined with established clinical predictors \cite{bib77,bib78,bib76}. Our findings support this latter body of evidence, suggesting that pretreatment DCE-MRI in isolation may not provide sufficient discriminative information for robust and generalizable pCR prediction across heterogeneous clinical centers.

At the same time, recent work has increasingly shifted toward longitudinal and multimodal strategies, integrating early on-treatment MRI, pathology, or tumor microenvironment information, which consistently report improved predictive performance \cite{bib79, bib80,bib81,bib82,bib83}. Together, these studies suggest that incorporating temporal and complementary biological information may be a more promising direction than relying solely on pretreatment imaging. We therefore view the MAMA-MIA benchmark not as evidence that pretreatment MRI has reached a definitive performance ceiling, but as a challenging and standardized baseline against which future methodological advances can be rigorously evaluated.

\subsection{On Subgroup Fairness as Challenge Objective}

Participating teams varied in how explicitly fairness was addressed during model development, with most submissions optimizing predictive performance and assessing fairness post-hoc through benchmark evaluation. A notable exception was the work from participant AIH-MAMA (FairMedSeg), which explicitly integrated subgroup fairness into their segmentation objective by penalizing performance gaps between subgroups during training. Other teams such as FME and MIC adopted robustness-oriented strategies, such as subtraction-based inputs or large-scale self-supervised pretraining and augmentation, thereby indirectly promoting more consistent performance across subgroups. 

Considering the pCR prediction, the results indicate that fairness considerations are meaningful primarily once a minimum level of predictive performance is achieved. As illustrated in Fig. \ref{fig1b}, low-performing models may appear particularly fair due to uniformly low predictions across subgroups, whereas fairness becomes a relevant and discriminative criterion only among models with competitive performance. This suggests that future benchmark editions are to consider adaptive weighting schemes in which fairness is emphasized more strongly the higher the performance of a downstream task model, thus aligning fairness incentives with clinical utility.

In this regard, the impact of the fairness weighting parameter $\lambda$ further highlights the normative nature of weighting the importance of subgroup fairness, particularly when dealing with tasks of differing clinical maturity. Our choice of an equal baseline weight ($\lambda = 0.5$) serves as a neutral starting heuristic; however, a fixed linear composite score has clear limitations. As observed in the lower-performing classification task, equal weighting risks inadvertently favoring fair but clinically uninformative models when predictive accuracy is close to chance.

These findings emphasize that fairness-aware benchmarking does not yield a single "correct" ranking, but instead exposes trade-offs that depend on institutional values, regulatory expectations, and clinical risk tolerance. Future iterations of such benchmarks should consider alternative formulation paradigms, such as performance-gated weighting (where a model must pass a baseline accuracy threshold to be eligible for fairness evaluation) or adaptive weighting schemes that scale $\lambda$ dynamically based on baseline task difficulty. By making these trade-offs explicit and adjustable, MAMA-MIA provides a transparent framework for interrogating how fairness objectives interact with performance.

We further note limitations of basing subgroup fairness on the differences between the means of performance metrics, as averaging subgroup performance can obscure relevant variability. For instance, models with similar mean fairness scores can exhibit substantially different, and likely undesirable, variability across subgroups. Moreover, mean-based across subgroups gaps can, in principle, be reduced by degrading performance in the highest-performing subgroup rather than by improving underperforming ones. This motivates future consideration of incorporating maximum-performance-based and variance–based measures to provide complementary insight into the robustness and reliability of reported fairness measures, particularly under heterogeneous subgroup sizes or outcome prevalence. 

\subsection{Implications for Future Benchmark Development and Clinical Translation}

The MAMA-MIA results highlight several directions for future benchmark design and model development. Most notably, the contrast between the strong performance achieved in primary tumor segmentation and the limited success of pretreatment pCR prediction suggests that current breast MRI AI has reached different levels of maturity across tasks.

While achieving impressive performance, the majority of submissions for primary tumor segmentation relied heavily on established convolutional neural networks (CNNs), particularly the self-configuring nnUNet framework \cite{bib13} and its variants. There remains substantial scope to investigate how these techniques can be modified, or how novel architectures \cite{bib84} can be introduced, to address specific vulnerabilities identified in this challenge, such as small tumors and the luminal molecular subtype. Ultimately, optimally tuning these newer approaches for multi-phase 3D DCE-MRI will be necessary to break through the performance ceilings established by the current CNN baselines.

For predicting pCR, participating methods exhibited substantial diversity in modelling architectures, pretraining strategies, and pre- and post-processing pipelines. However, overall performance remained limited, with no method exceeding an AUC of ~0.57. This suggests that the challenge setting, characterized by domain shift, remains difficult to address using imaging information alone.
The results further indicate that future progress will likely require the integration of treatment-specific variables (e.g., regimen, dosage), molecular subtype, and proliferation markers (e.g., Ki-67 \cite{bib85}) to contextualize the imaging findings. In particular, the observed performance variability across molecular subtypes and patient subgroups indicates that response prediction cannot be treated solely as an imaging problem, but must account for the underlying biological heterogeneity of breast cancer.
Likewise, extending evaluation beyond binary pCR classification toward more continuous measures of treatment response, such as residual disease burden or tumor volume change, may provide richer information and better reflect clinically relevant treatment outcomes. The inclusion of longitudinal imaging acquired during therapy could further help capture treatment-induced changes that are not observable from pretreatment imaging alone \cite{bib86, bib87}.

From a benchmarking perspective, the challenge also revealed several opportunities for methodological refinement. The sensitivity of leaderboard rankings to the fairness weighting parameter $\lambda$ demonstrates the importance of transparently reporting performance–fairness trade-offs and motivates future investigation of adaptive or performance-gated fairness formulations. Similarly, computational efficiency was not explicitly included in the challenge ranking, despite its relevance to clinical usability. While a strict upper limit of 5 minutes per case was defined, individual runtime measurements were not considered sufficiently reproducible for leaderboard ranking, as execution times in containerized benchmark environments can be affected by implementation details and infrastructure-related variability. Future benchmark editions could investigate standardized efficiency metrics, including inference latency, memory consumption, and energy usage, measured under controlled execution settings.

\section{Conclusion}

This work introduced the MAMA-MIA benchmark, a large-scale, multi-center framework designed to jointly evaluate generalization and subgroup fairness in breast MRI tumor segmentation and pretreatment pCR prediction. While automated tumor segmentation demonstrated robust performance and consistent generalization across unseen clinical sites, pCR prediction from pretreatment DCE-MRI showed only marginal gains over baseline under realistic cross-center evaluation.

These findings suggest that, despite encouraging progress reported in recent imaging-based pCR prediction studies, substantial challenges remain in achieving robust and reproducible performance under a common multi-center evaluation framework. By providing standardized evaluation across institutions, countries, imaging protocols, and clinically relevant patient subgroups, MAMA-MIA offers a complementary perspective to individual model-development studies and helps quantify the gap between experimental performance and real-world generalizability.

More broadly, the challenge demonstrates the value of integrating subgroup fairness alongside predictive performance when assessing medical AI systems. We hope that MAMA-MIA will serve as a lasting benchmark resource for the community, supporting the development of breast MRI models that are not only accurate, but also robust, equitable, and clinically deployable across diverse patient populations.

\section*{Acknowledgment}

This project has received funding from the European Union’s Horizon 2020 research and innovation programmes under grant agreement No 952103 (EUCanImage) and No 101057699 (RadioVal). Also, this work was partially supported by the project FUTURE-ES (PID2021-126724OB-I00) and project AIMED (PID2023-146786OB-I00) from the Ministry of Science and Innovation of Spain. K.K. holds a RyC fellowship grant (RYC2024-048256-I).

Team Martel Lab (M. A., T.X., and A.M.) are supported by the Canada Foundation for Innovation (40206), and the Ontario Research Fund. Computational resources were provided in part by the Digital Research Alliance of Canada (alliancecan.ca). Team ViCOROB (H.A., J.C.V. and R.M.) funded by the projects VICTORIA, “PID2021-123390OB-C21” and IMPACT, “PID2024-157201OB-C21” from the Ministerio de Ciencia, Innovación y Universidades of Spain, and the PhD grant IFUdG2024 from the University of Girona. Team pimed-lab (D.S., J.H.L. and M.R.) is supported by the National Cancer Institute of the National Institutes of Health under Award Number R37CA260346. Team AIH-MAMA (E.P., L.V., E.P., M.A.Z.) is partly supported by the ANR-BMBF TRAIN (ANR-22-FAI1-0003-02). We would like to thank all team members who helped during the benchmark: Daniele Falcetta (AIH-MAMA), Vincenzo Marciano' (AIH-MAMA), Tania Cerquitelli (AIH-MAMA), Elena Baralis (AIH-MAMA).

\bibliographystyle{IEEEtran}
\bibliography{references}

\end{document}